\documentclass[preprint,authoryear,table, xcdraw, 12pt]{elsarticle} 

\usepackage{graphics}
\usepackage{graphics}
\usepackage{float}
\usepackage{graphicx}
\usepackage{wrapfig}

\usepackage[labelfont=bf]{caption}
\usepackage{adjustbox}
\usepackage{caption}

\usepackage{pbox}

\usepackage{subcaption}
\usepackage{color}
\usepackage{amsmath} % Mathe
\usepackage{mathtools} % Mathe
\usepackage{amsfonts} % Mathesymbole
\usepackage{amssymb}

\usepackage{array}
\usepackage{graphicx}
\usepackage[graphicx]{realboxes}

\usepackage{booktabs}
\usepackage{rotating}
\usepackage{multirow}
\usepackage{multicol}
\usepackage{lscape}

\usepackage{xargs}                      % Use more than one optional parameter in a new commands
\usepackage{calc}

\usepackage[draft,colorinlistoftodos,prependcaption,textsize=tiny]{todonotes}

\usepackage{url}

\newcommandx{\ISLEM}[2][1=]{\todo[linecolor=blue,backgroundcolor=blue!25,bordercolor=blue,#1]{#2}}

\newcommandx{\ISMAIL}[2][1=]{\todo[linecolor=green,backgroundcolor=green!25,bordercolor=green,#1]{#2}}

\usepackage{color}
\usepackage{xcolor}
\usepackage{colortbl}

\usepackage{soul}

\definecolor{babyblue}{rgb}{0.54, 0.81, 0.94}
\definecolor{armygreen}{rgb}{0.29, 0.33, 0.13}
\definecolor{brightlavender}{rgb}{0.75, 0.58, 0.89}
\definecolor{aqua}{rgb}{0.0, 1.0, 1.0}
\definecolor{caribbeangreen}{rgb}{0.0, 0.8, 0.6}
\definecolor{reddish}{rgb}{0.82, 0.1, 0.26}
\definecolor{caribbeangreen}{rgb}{0.31, 0.78, 0.47}
\definecolor{jasper}{rgb}{0.84, 0.23, 0.24}
\definecolor{red}{rgb}{1.0, 0.0, 0.0}
\definecolor{green}{rgb}{0.0, 1.0, 0.0}
\definecolor{blue}{rgb}{0.0, 0.0, 1.0}
\definecolor{darkgreen}{rgb}{0.1, 0.7, 0.1}
\definecolor{darkblue}{rgb}{0.1, 0.1, 0.7}
\definecolor{red}{rgb}{0.7, 0.1, 0.1}
\definecolor{coral}{rgb}{1.0, 0.5, 0.31}

\definecolor{babypink}{rgb}{0.96, 0.76, 0.76}
\definecolor{bananamania}{rgb}{0.98, 0.91, 0.71}
\definecolor{bananayellow}{rgb}{1.0, 0.88, 0.21}
\definecolor{maize}{rgb}{0.98, 0.93, 0.37}
\definecolor{champagne}{rgb}{0.97, 0.91, 0.81}

\usepackage[colorlinks=true,linkcolor=coral,citecolor=coral]{hyperref}
\usepackage[colorinlistoftodos,prependcaption,textsize=tiny]{todonotes}

\definecolor{darkgreen}{rgb}{0.53, 0.66, 0.42}

\usepackage[ruled,vlined]{algorithm2e}

\journal{}

\begin{document}

\begin{frontmatter}



%Multi-view Multi-layer
%\title{Computer-Aided Diagnosis Methods in Autism Spectrum Disorder Classification Kaggle Challenge using Cortical Morphological Networks}

\title{A Comparative Study of Machine Learning Methods for Predicting the Evolution of Brain Connectivity from a Baseline Timepoint}

%A Python Toolbox for Machine Learning Brain Classification

%\author[BASIRA]{Anna Lisowska}\right) 
\author{\c{S}eymanur Akt{\i}$^{\dagger}$ \fnref{ITU}}
\author{Do\u{g}ay Kamar$^{\dagger}$ \fnref{ITU}}
\author{\"{O}zg\"{u}r An{\i}l \"{O}zl\"{u} \fnref{ITU}}
\author{Ihsan Soydemir \fnref{ITU}}
\author{Muhammet Akcan \fnref{ITU}}
\author{Abdullah Kul \fnref{ITU}}
\author{Islem Rekik \corref{cor} \fnref{BASIRA,DUNDEE}}

\address[ITU]{Faculty of Computer and Informatics, Istanbul Technical University}
\address[BASIRA]{BASIRA lab, Faculty of Computer and Informatics, Istanbul Technical University, Istanbul, Turkey}
\address[DUNDEE]{School of Science and Engineering, Computing, University of Dundee, UK}

\cortext[same]{$^{\dagger}$ Co-first authors: akti15@itu.edu.tr and kamard@itu.edu.tr}

\cortext[cor]{Corresponding author; Dr Islem Rekik (irekik@itu.edu.tr), \url{http://basira-lab.com/}, GitHub code: \url{https://github.com/basiralab/Kaggle-BrainNetPrediction-Toolbox}} %brain connectome diganosis BCD-kit


%------------------------------------------------------------------------------------------------%

% Checklist
% Put the abbrv. in first place they are used e.g ASD, NC, fMRI, dMRI

\begin{abstract}

Predicting the evolution of the brain network, also called connectome, by foreseeing changes in the connectivity weights linking pairs of anatomical regions makes it possible to spot connectivity-related neurological disorders in earlier stages and detect the development of potential connectomic anomalies. Remarkably, such a challenging prediction problem remains least explored in the \emph{predictive connectomics} literature. It is a known fact that machine learning (ML) methods have proven their predictive abilities in a wide variety of computer vision problems. However, ML techniques specifically tailored for the prediction of brain connectivity evolution trajectory from a single timepoint are almost absent. To fill this gap, we organized a Kaggle competition where 20 competing teams designed advanced machine learning pipelines for predicting the brain connectivity evolution from a single timepoint. The teams developed their ML pipelines with combination of data pre-processing, dimensionality reduction and learning methods. Each ML framework inputs a baseline brain connectivity matrix observed at baseline timepoint $t_0$ and outputs the brain connectivity map at a follow-up timepoint $t_1$. The longitudinal OASIS-2 dataset was used for model training and evaluation. Both random data split and 5-fold cross-validation strategies were used for ranking and evaluating the generalizability and scalability of each competing ML pipeline. Utilizing an inclusive approach, we ranked the methods based on two complementary evaluation metrics (mean absolute error (MAE) and Pearson Correlation Coefficient (PCC)) and their performances using different training and testing data perturbation strategies (single random split and cross-validation). The final rank was calculated using the rank product for each competing team across all evaluation measures and validation strategies. Furthermore, we added statistical significance values to each proposed pipeline. In support of open science, the developed 20 ML pipelines along with the connectomic dataset are made available on GitHub\footnote{\url{https://github.com/basiralab/Kaggle-BrainNetPrediction-Toolbox}}. The outcomes of this competition are anticipated to lead the further development of predictive models that can foresee the evolution of the brain connectivity over time, as well as other types of networks (e.g., genetic networks).

%However, further improvements are still needed to reach the clinical translation of functional connectomics. We have kept the CNI-TLC open as a publicly available resource for developing and validating new classification methodologies in the field of predictive connectomics.

%After analyzing the pipelines, it is concluded that the diversity of chosen methods were sufficient and the results embody the competence of machine learning methods in solving the estimation of brain connectivity map. Therefore, the outcome of this competition is anticipated to lead the further research on the estimation of brain connectivity from a single observation.

\end{abstract}


\end{frontmatter}

%%%%%%%%%%%%%%%%%%%%%%%%%%%%%%%%%%%%%%%%%%%%%%%%%%%%%%%%%%%%%%%%%%%%%%%%%%%%%%%%%%
\section{Introduction} 
%%%%%%%%%%%%%%%%%%%%%%%%%%%%%%%%%%%%%%%%%%%%%%%%%%%%%%%%%%%%%%%%%%%%%%%%%%%%%%%%%%

\begin{figure}[h!]
    \centering
    \includegraphics[width=14cm]{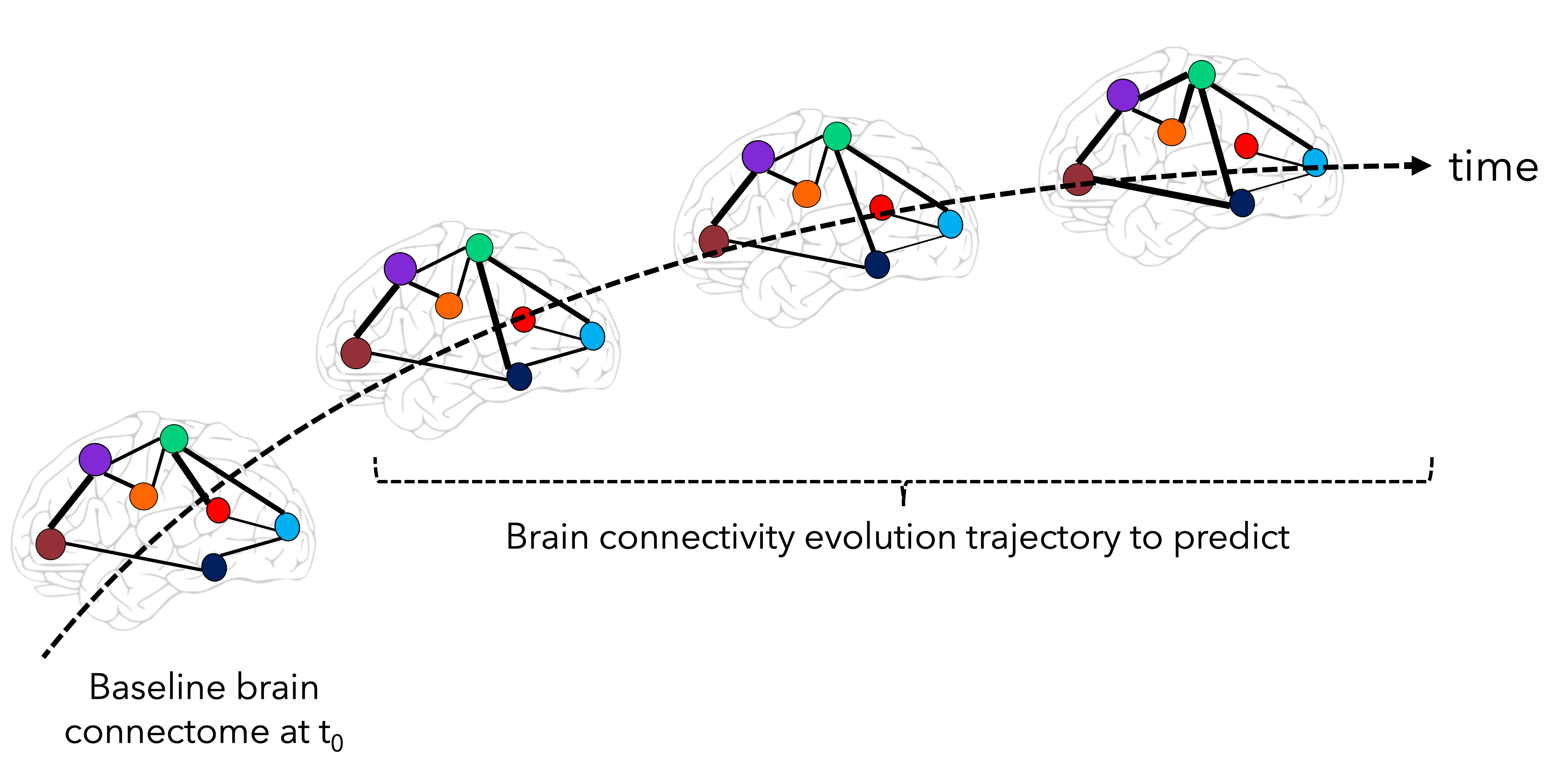}
    \caption{\emph{Predicting the evolution trajectory of brain connectivity from a baseline timepoint $t_0$.} In this illustration, we sketch the scenario where the brain connectivity weights between 7 anatomical regions of interest (ROIs) change over time. For instance, brain connectivity weakening can be possibly explained by typical neural pruning during postnatal development or cognitive decline during typical aging.}
    \label{fig:concept}
\end{figure}

% * The problem statement and motivation. Why do we need brain connectivity prediction?
%	- The problem is predicting the brain connectivity map for a further time step given the current brain connectivity map.
%	- For which problems we can benefit from brain connectivity prediction task? (Predicting the possible diseases beforehand, detecting anomalies on brain evolution etc.)
The brain connectome presents a compact representation of interactions between different anatomical regions of interest (ROIs) using non-invasive magnetic resonance imaging (MRI) \citep{zalesky2012connectivity,fornito2015connectomics,van2019cross}. In fact, the connectome unravels the complexity of the brain as a highly inter-connected system by modeling its multi-type interactions such as correlation in neural firing \citep{park2013structural,power2010development,tomasi2012aging} and similarity in morphology \citep{soussia2018,corps2019,dhifallah2020} or fiber structure \citep{khundrakpam2013developmental,wen2011structural,bullmore2009complex}. The study of the human brain connectivity has attracted unparalleled attention in the last decades enabling to chart out a cross-disorder landscape of brain disconnectivity such as Alzheimer's disease, dementia, and schizophrenia among other disorders \citep{van2019cross}. Several studies showed that there is a wide spectrum of valuable information which can be extracted from the brain connectivity such as human behaviour \citep{shen2017using}, individual creative ability \citep{beaty2018robust}, Alzheimer's disease \citep{khazaee2016application} and autism \citep{deshpande2013identification} biomarkers, and way more. More importantly, tracking the evolution of the brain connectome, where connectivity weights between pairs of ROIs evolve over time, allows us to map the dynamics of both typical and atypical alterations in the brain as a network \citep{serra2016longitudinal,ghribi2019progressive,filippi2020longitudinal} (\textbf{Fig.}~\ref{fig:concept}). The temporal evolving nature of the connectome presents a formidable challenge for \emph{predictive intelligence models} \citep{soussia2018review}, however bringing with it the unprecedented opportunity to foresee the time-dependent changes in the brain connectivities over time. Such predictive models can probe the early diagnosis of connectivity-related neurological disorders and predict eventual anomalies from a single timepoint (i.e., one baseline MRI). Thus, given that the predictive ability of machine learning methods has remarkably improved over the last years, such intelligent predictive models can be deployed for predicting the brain connectome evolution over time from minimal data (i.e., a single baseline connectome). 

%* Reporting some related works
%	Deep learning methods
%	- Weakness: Requires high computation power
%	- Our strength: Getting promising results even with machine learning methods
%	Non-deep-learning methods
%	- Weakness: Only a small portion of the methods are analysed.
%	- Our strength: A wide variety of machine learning methods are employed to test their abilities on the task.

A few pioneering works aimed to predict the brain evolution from a single timepoint however by using an image-based \citep{rekik2015prediction,gafurouglu2018joint} or surface-based representations \citep{rekik2016predicting,rekik2017joint}. Notably, both representations discard the \emph{connectional} aspect of the brain as a network. Recently, very few works aimed to predict the evolution of the brain connectivity from a single observation in infant \citep{ghribi2019progressive} and elderly populations \citep{ezzine2019learning,gurler2020foreseeing,nebli2020deep}. For instance, \citep{ghribi2019progressive} introduced a supervised sample-based selection strategy for learning how to select the best training samples for predicting the baby connectome trajectory evolution from a neonatal connectivity map. \citep{ezzine2019learning} modeled the subject-specific evolution trajectory of a brain connectome as a learnable deviation from the average population-driven brain connectome (called network atlas). \citep{gurler2020foreseeing, nebli2020deep} exploited deep learning methods where they used graph-based generative adversarial networks to predict the longitudinal brain connectivity from a single timepoint. Even though these methods can predict the brain evolution trajectory with satisfying accuracy, the deep learning oriented methods require high computation resources with complex architectures to train and hyper-parameters to tune. Furthermore, so far existing competitions using connectomic data focused on brain connectivity classification such as the Kaggle 2019 competition\footnote{\url{https://github.com/basiralab/BrainNet-ML-ToolBox}} including 20 teams for autism spectrum disorder diagnosis \citep{BILGEN2020108799} and the MICCAI Connectomics in Neuroimaging Transfer Learning Challenge\footnote{\url{https://github.com/aichung/pl-cni_challenge}} including 5 teams \citep{schirmer2021} for neuropsychiatric disorder diagnosis. In this paper, we present and analyze cutting-edge machine learning pipelines designed in an in-class  Kaggle 2021 competition for predicting the temporal changes in brain connectivity from a single timepoint, offering lighter and simpler models which can run faster and be easily deployed in clinical facilities.

%* Emphasizing our competition's strength and contribution
%	The brain connectivity map is predicted using only brain maps and wide range of methods are experimented.
%There is a big gap in the literature around this topic and it still remains to be explored. Therefore, an in-class Kaggle competition is organized and teams utilized a wide range of machine learning methods to solve the problem of brain connectivity prediction in a further time step given a single observation. This being a formidable problem, the results indicate the capability of machine learning approaches on this specific task. 

%* Details about competition
%	Teams tackled the brain connectivity prediction task on the given dataset.
%	The methods from top 20 teams are ranked considering PCC and MAE results (statistical analyses). MEDIA 2020 (violin)

For the in-class Kaggle challenge, each team constructed their own pipeline including the pre-processing and training steps. The teams used  morphological brain connectivity maps from the OASIS-2 dataset where they were given the connectivity maps at $t_0$ and expected to predict the connectivity map at a 1-year follow-up timepoint $t_1$ of each individual. The designed ML frameworks include but are not limited to linear regression, support vector regression, ridge regression, decision trees and ensemble methods. We ranked all competing teams based on two complementary evaluation criteria: the mean absolute error (MAE) and Pearson correlation coefficient (PCC) scores. Keeping the metrics standardized, the prediction results fairly evaluate the performance of each method. In addition, we have presented some statistical analyses to compare the proposed approaches in depth. For reproducibility, the longitudinal morphological connectomic dataset along with the designed ML pipeline codes are made available at \url{https://github.com/basiralab/Kaggle-BrainNetPrediction-Toolbox}.

%* Summary of the paper
%	Methods: the method of each team will be presented
%	Results: Results are analysed in results section
%	Discussion: Weaknesses and further improvements are mentioned at 
%	Conclusion

In the following sections of the paper, we present the designed ML pipelines by each team and stress their motivation on using the chosen components, include an extensive analysis of overall results and discuss the shortcomings of the competition outcomes with mention of potential further improvements.

%%%%%%%%%%%%%%%%%%%%%%%%%%%%%%%%%%%%%%%%%%%%%%%%%%%%%%%%%%%%%%%%%%%%%%%%%%%%%%%%%%
\section{Methodology}
%%%%%%%%%%%%%%%%%%%%%%%%%%%%%%%%%%%%%%%%%%%%%%%%%%%%%%%%%%%%%%%%%%%%%%%%%%%%%%%%%%

In this section, we detail the competition organization, introduce the predictive machine learning frameworks proposed by the competing teams and compare their performances for the temporal brain connectivity prediction task.

\subsection{Competition Organization}

The brain evolution estimation using connectivity maps can be formalized as a regression problem where the proposed model is trained with the initial brain connectivity map at $t_0$ and expected to predict the connectivity map at a follow-up timepoint $t_1$. The type of data referred to as `brain connectivity map' is basically a matrix quantifying the interactions between pairs of anatomical regions of interest (\textbf{Fig.}~\ref{fig:concept}). The matrix presents the adjacency matrix of a brain graph where nodes present ROIs and edge weights present the score assigned to their interaction type (e.g., Pearson correlation for synchrony in neural activity). Learning such as time-dependent mapping with a generalized approach allows the trained ML model to predict future connectivity maps from unseen brain connectomes acquired at baseline $t_0$ in the testing phase. Hence, one can foresee the evolution of the brain connectivity at follow-up timepoints based on a \emph{single} observation. 

To investigate the efficacy of a diverse pool of ML pipelines in predicting the evolution trajectory of the brain connectome, an in-class Kaggle competition is organized using the OASIS-2 dataset \citep{rahman2018oasis} where we derived cortical morphological networks \citep{soussia2018,nebli2020} from T1-weighted MRI at two timepoints, 1-year apart. We split the dataset into training and test sets and we instructed teams to train their models using a public training set including both $t_0$ and $t_1$ morphological connectomes. After completing the training, they tested their models using a test set including connectivity maps at $t_0$ maps. They submitted their predicted connectivity maps at follow-up timepoint $t_1$ through the Kaggle system and the results were compared with the ground truth $t_1$ maps to evaluate their pipelines. During the competition, the teams were able to check their results only on half of the test set (i.e., public ranking) which provided them with a publicly shared performance score while the other half of the test set was kept private for extra evaluation. The private performance scores were only revealed after the Kaggle competition ended. %Furthermore, the number of submissions per day was limited, hence the results are calculated fairly by forcing the proposed ML pipelines to generalize to unseen samples instead of memorizing the test set. %The overall test set results are evaluated and presented after the competition ends. 

Teams were ranked based on two evaluation measures (MAE and PCC). In addition to the public and private rankings, all teams evaluated their models by using 5-fold (5F) cross-validation on the shared training set. This assesses the generalizability and stability of the proposed pipeline against data distribution perturbations. For a more robust and conclusive assessment of each designed pipeline, we ranked the competing teams based on their combined results of MAE and PCC scores on the public test set, private test set and 5F cross-validation. This ensures that the highly ranked models are most \emph{reproducible} (i.e., the model performance is stable against varying training and testing sets) and \emph{scalable} (i.e., the model performance is stable against varying scales of the training set).

\subsection{Analysis of the Proposed Machine Learning Pipelines for Brain Connectivity Prediction}

The tailored machine learning pipelines are composed of different blocks such as data processing and dimensionality reduction. Here we grouped the blocks into three categories: (1) pre-processing methods, (2) dimensionality reduction methods and (3) learning models. In this section, we present and analyze the pipelines of the top 20 teams in each of the 3 categories. \textbf{Table}~ \ref{tab:method_component} provides an overview on the used components by each team where the teams are sorted based on their final ranks.

\renewcommand{\arraystretch}{1.8}
%\Rotatebox{90}{
\begin{table}[h!]
  \centering
  \resizebox{1.0\textwidth}{!}{%
 % \tiny
\begin{tabular}{|l|>{\columncolor[HTML]{f4e9dc}}c|>{\columncolor[HTML]{f4e9dc}}c|>{\columncolor[HTML]{f4e9dc}}c|>{\columncolor[HTML]{f4e9dc}}c|>{\columncolor[HTML]{f4e9dc}}c|>{\columncolor[HTML]{f4e9dc}}c|>{\columncolor[HTML]{f4e9dc}}c|>{\columncolor[HTML]{f4e9dc}}c|>{\columncolor[HTML]{f4e9dc}}c|>{\columncolor[HTML]{f4e9dc}}c|>{\columncolor[HTML]{d8e3e3}}c|>{\columncolor[HTML]{d8e3e3}}c|>{\columncolor[HTML]{d8e3e3}}c|>{\columncolor[HTML]{d8e3e3}}c|>{\columncolor[HTML]{d8e3e3}}c|>{\columncolor[HTML]{d8e3e3}}c|>{\columncolor[HTML]{f7edec}}c|>{\columncolor[HTML]{f7edec}}c|>{\columncolor[HTML]{f7edec}}c|>{\columncolor[HTML]{f7edec}}c|>{\columncolor[HTML]{f7edec}}c|>{\columncolor[HTML]{f7edec}}c|>{\columncolor[HTML]{f7edec}}c|>{\columncolor[HTML]{f7edec}}c|>{\columncolor[HTML]{f7edec}}c|>{\columncolor[HTML]{f7edec}}c|>{\columncolor[HTML]{f7edec}}c|>{\columncolor[HTML]{f7edec}}c|>{\columncolor[HTML]{f7edec}}c|>{\columncolor[HTML]{f7edec}}c|>{\columncolor[HTML]{f7edec}}c|>{\columncolor[HTML]{f7edec}}c|>{\columncolor[HTML]{f7edec}}c|>{\columncolor[HTML]{f7edec}}c|>{\columncolor[HTML]{f7edec}}c|>{\columncolor[HTML]{f7edec}}c|}
\hline
%ANoise	OE	LOF	IF	CFE	InvSF	Scaler	IQR	CorB	RD	GUS	PCA	VT	SKB	TrSVD	BE	BR	HR	Ridge	LGMBReg	KNReg	Enet	GBReg	ABReg	Lasso	OMP	XGBReg	Lreg	RFR	BagReg	SVR	KMClus	SNReg	DTR	ERTree	FFL
\textbf{Teams} & \textbf{ANoise} & \textbf{OE} & \textbf{LOF} & \textbf{IF} & \textbf{CFE} & \textbf{InvSF} & \textbf{Scaler} & \textbf{IQR} & \textbf{CorB} & \textbf{RD} & \textbf{GUS} & \textbf{PCA} & \textbf{VT} & \textbf{SKB} & \textbf{TrSVD} & \textbf{BE} & \textbf{BR} & \textbf{HR} & \textbf{Ridge} & \textbf{LGMBReg} & \textbf{KNReg} & \textbf{Enet} & \textbf{GBReg} & \textbf{ABReg} & \textbf{Lasso} & \textbf{OMP} & \textbf{XGBReg} & \textbf{LReg} & \textbf{RFR} & \textbf{BagReg} & \textbf{SVR} & \textbf{KMClus} & \textbf{SNReg} & \textbf{VotingR} & \textbf{ERTree} & \textbf{FFL} \\ \hline
Team-1 & & & \cellcolor[HTML]{F4BF81} & & & & & & & & & & & & & & \cellcolor[HTML]{F49C92} & & & & & & & & & & & & & & & & & & & \cellcolor[HTML]{F49C92}\\ \hline
Team-2 & & & & & & & & & & & & & & & & & & \cellcolor[HTML]{F49C92}& & & & & & & & & & & & & & & & & & \cellcolor[HTML]{F49C92}\\ \hline
Team-11 & & & & & & & & \cellcolor[HTML]{F4BF81} & & & & & & & & & & & & & & & & & & & & \cellcolor[HTML]{F49C92} & & & & & & & & \cellcolor[HTML]{F49C92}\\ \hline
Team-13 & & \cellcolor[HTML]{F4BF81} & & & & & & & & & & & & & & & & & & & & & & & & & & \cellcolor[HTML]{F49C92}  & & & & & & & &\\ \hline

Team-3 & & & & & & & & & & & \cellcolor[HTML]{68CBD0} & \cellcolor[HTML]{68CBD0} & & & & & & & \cellcolor[HTML]{F49C92} & \cellcolor[HTML]{F49C92} & \cellcolor[HTML]{F49C92} & \cellcolor[HTML]{F49C92} & \cellcolor[HTML]{F49C92} & \cellcolor[HTML]{F49C92} & \cellcolor[HTML]{F49C92} & \cellcolor[HTML]{F49C92} & \cellcolor[HTML]{F49C92} & & & & & & &\cellcolor[HTML]{F49C92} & &\cellcolor[HTML]{F49C92}\\ \hline
Team-12 & & & & & & & & & & & & & & & & & & & & & & & & & & & & & & & & & \cellcolor[HTML]{F49C92} & & &\cellcolor[HTML]{F49C92} \\ \hline
Team-7 & & & & & & \cellcolor[HTML]{F4BF81}  & & & & & & & & \cellcolor[HTML]{68CBD0} & & & & & & & & & & & & & & & & & \cellcolor[HTML]{F49C92} & & & & &\\ \hline
Team-19 & & & & & & \cellcolor[HTML]{F4BF81} & & & & & & & & \cellcolor[HTML]{68CBD0} & & & & & & & & & & & & & & & & & \cellcolor[HTML]{F49C92} &  & & & &\\ \hline
Team-18 & \cellcolor[HTML]{F4BF81} & & & & & & & & & & & & & & & & & & & & & & & & & & & & & & & & & & \cellcolor[HTML]{F49C92} &\cellcolor[HTML]{F49C92} \\ \hline
Team-4 & & & & \cellcolor[HTML]{F4BF81} & & & \cellcolor[HTML]{F4BF81} & & & & & &  \cellcolor[HTML]{68CBD0} & & & & & & \cellcolor[HTML]{F49C92} & & & & & & & & & \cellcolor[HTML]{F49C92} & & & & &  & \cellcolor[HTML]{F49C92} & &\\ \hline
Team-16 & & & & & \cellcolor[HTML]{F4BF81} & & & & & \cellcolor[HTML]{F4BF81}  & & & & & & & & & & & & & & \cellcolor[HTML]{F49C92}  & & & & & & & & & & & &\\ \hline

Team-15 & & & & & & & & & & \cellcolor[HTML]{F4BF81} & & & & & & & \cellcolor[HTML]{F49C92} & & & & \cellcolor[HTML]{F49C92} & & & \cellcolor[HTML]{F49C92} & & & & & & & & & & \cellcolor[HTML]{F49C92} & &\\ \hline
Team-5 & & \cellcolor[HTML]{F4BF81} & & & & & & & & & & & & & & & & & & & & & & & & & & & \cellcolor[HTML]{F49C92} & & & & & & &\\ \hline
Team-17 & & \cellcolor[HTML]{F4BF81} & & & & & \cellcolor[HTML]{F4BF81} & & \cellcolor[HTML]{F4BF81} & & & & & & & \cellcolor[HTML]{68CBD0} & & & \cellcolor[HTML]{F49C92} & & & & & & & & & & & & & & & & &\\ \hline
Team-20 & & & & & \cellcolor[HTML]{F4BF81} & & & & & \cellcolor[HTML]{F4BF81} & & \cellcolor[HTML]{68CBD0} & & & & & & & & & & & & & &  & & & & & \cellcolor[HTML]{F49C92} & & & & &\\ \hline
Team-9 & & & & & & & & & & & & & & & \cellcolor[HTML]{68CBD0} & & & & & & & & & & & & & & \cellcolor[HTML]{F49C92} & & & \cellcolor[HTML]{F49C92} &  & \cellcolor[HTML]{F49C92} & &\\ \hline
Team-8 & & & & & \cellcolor[HTML]{F4BF81}  & & & & & & & & & \cellcolor[HTML]{68CBD0} & & & & & & & \cellcolor[HTML]{F49C92} & & \cellcolor[HTML]{F49C92} & \cellcolor[HTML]{F49C92} & & & & & & & & & & \cellcolor[HTML]{F49C92} & &\\ \hline
Team-14 & & & & \cellcolor[HTML]{F4BF81} & \cellcolor[HTML]{F4BF81} & & & & & & & & & & & & & & \cellcolor[HTML]{F49C92} & & & & & & & & & & & & & & & & &\\ \hline

Team-6 & & & \cellcolor[HTML]{F4BF81} & & \cellcolor[HTML]{F4BF81} & & & & & & & & & & & & & & & & & & & & & & & & & \cellcolor[HTML]{F49C92} & & & & & &\\ \hline
Team-10 & & & & \cellcolor[HTML]{F4BF81} & & & \cellcolor[HTML]{F4BF81} &  & & & & \cellcolor[HTML]{68CBD0} & & & & & & & & & & & & & & & & & \cellcolor[HTML]{F49C92} & & & & & & &\\ \hline
\end{tabular}
}
\vspace{6pt}
\caption{\scriptsize Preprocessing (orange), dimensionality reduction (blue) and learning methods (red) used by the top 20 teams. The teams were sorted based on the final ranking. The method names are abbreviated as: Add Noise (ANoise), Outlier Elemination (OE), Local Outlier Factor (LOF), Isolation Forest (IF), Constant Feature Eleminaion (CFE), Inverse Sigmoid Function (InvSF), Correlation-based Feature Elimination (CorB), Redundant Feature Elimination (RD), Generic Univariate Select (GUS), Principal Component Analysis (PCA), Variance Threshold (VT), Select K Best(SKB), TruncatedSVD (TrSVD), Backward Elemination (BE), Bayesian Ridge Regressor (BR), Huber Regressor (HR), LGBMRegressor (LGMBReg), KNeighbors Regressor (KNReg), ElasticNet (Enet), Gradient Boosting Regressor (GBReg), AdaBoost Regressor (ABReg), Orthogonal Matching Pursuit (OMP), XGBRegressor (XGBReg), Linear Regression (LReg), Random Forest Regressor (RFE), Bagging Regressor (BagReg), Support Vector Regressor (SVR), Kmeans Clustering (KMClus), Singular Neighbors Regressor (SNReg), Extra Random Trees (ERTree), Feature Focused Learning (FFL).}
\label{tab:method_component}%
\end{table}

\subsubsection{Preprocessing Techniques}

Data pre-processing is a preliminary step to several machine learning pipelines, which can include data denoising and the removal of irrelevant samples or features. Such techniques produce better representations of the data in a more compact way and thus the learner can easily extract valuable information in the target learning task.

\begin{figure}[h!]
    \centering
    \includegraphics[width=0.9\textwidth]{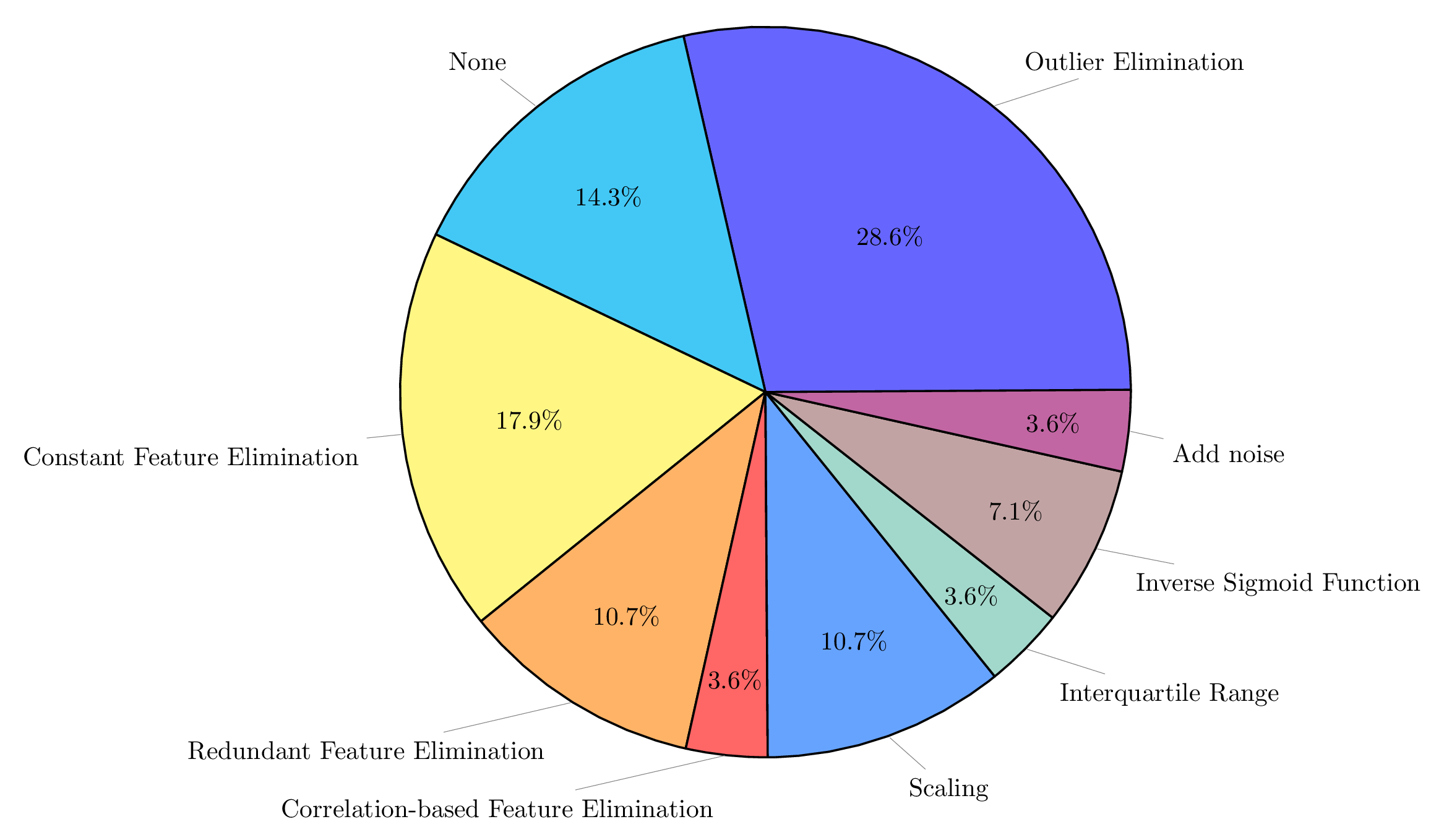}
    \caption{Distribution of chosen pre-processing techniques among the 20 competing teams.}
    \label{fig:preprocess}
\end{figure}

The pre-processing methods that the competing teams utilized prior to training their models include outlier elimination, noise adding, inverse sigmoid function, interquartile range, scaling, correlation-based feature elimination, redundant feature elimination and constant feature elimination. Four of the teams (Team 2, Team 3, Team 9 and Team 12) preferred not to use any pre-processing techniques on the data. \textbf{Table}~ \ref{tab:method_component} shows that a few teams were able to get relatively higher ranks than the others without any pre-processing step, indicating that the pre-processing may not be a mandatory block in the ML pipeline for the target prediction task. The distribution of the selected pre-processing techniques among teams is displayed in \textbf{Fig.}~\ref{fig:preprocess}.

The commonly used pre-processing techniques among teams are outlier elimination methods where three teams (Team 5, Team 13, Team 17) used their own approach to eliminate outliers while Team 11 used Inter-quartile Range Method, two teams (Team 1, Team 6) used Local Outlier Factor and three teams (Team 4, Team 10, Team 14) used Isolation Forest. These outlier elimination methods eliminate the out-of-distribution samples preventing the model to be misled by them.

The custom outlier elimination methods are (1) eliminating the samples that lay outside of the $ \mu \pm 3 \times \sigma$ ($\mu$ denotes the distribution average and $\sigma$ its standard deviation)  interval considering the normalized data distribution (used by Team 5 and Team 13) and (2) training the model by removing samples one by one and getting the optimum sample set (used by Team 17). Outlier elimination is the only pre-processing method that Team 5 and Team 13 used as both have simpler pipelines. Team 17 used outlier elimination along with other pre-processing techniques such as scaling, correlation based feature elimination and backward feature elimination. 

For the Inter-quartile Range Method, the samples that fall outside of the lower and upper bounds are detected as outliers. The lower and upper bounds are determined as shown at equation \ref{eq:IQR1} and \ref{eq:IQR2}.
\begin{equation}
    \centering
    Upper Bound = Q3 + 1.5 \times IQR 
    \label{eq:IQR1}
\end{equation}
\begin{equation}
    \centering
    Lower Bound = Q1 - 1.5 \times IQR
    \label{eq:IQR2}
\end{equation}

where IQR is the Inter-quartile range calculated as $IQR = Q3 - Q1$. Here $Q1$ denotes the first quartile of given range and $Q3$ denotes the third quartile.

Inter-quartile Range  method is the only pre-processing method that Team 11 used together with Linear Regressor and feature focused learning. Note that Team 11 ranked 3rd among the 20 competing teams. 

Local Outlier Factor \citep{breunig2000lof} method is an unsupervised anomaly (i.e outlier) detection method that assigns an anomaly score to each sample by comparing their local densities with the local densities of its neighbors. If there is a big difference between the local densities, this sample can be considered as an isolated sample and likely to have a higher anomaly score which makes it an outlier. The other method used for outlier elimination, Isolation Forest \citep{liu2008isolation}, assigns anomaly scores to data samples based on their average depths in a group of trees where the trees are created by splitting the range of features recursively. Team 14 used Isolation Forest together with Constant Feature Elimination and did not use any other dimensionality reduction techniques similar to Team 6. Teams 4 and 10 both used scaler as a pre-processing step along with Isolation Forest and further used dimensionality reduction methods. 
%When the rankings of the teams that used this method is compared at Table \ref{tab:method_component}, Team 4 ranked higher than the Team 10 and Team 14. When we consider that the pre-processing methods of Team 4 and Team 10 are really similar, it can be again addressed that the learning model plays the key role on the performance comparison of these 2 teams. Team 14 also used Constant Feature Elimination together with outlier elimination as Team 6 did.

Constant Feature Elimination (CFE) is the second most popular method as it can be seen at \textbf{Fig.}~\ref{fig:preprocess} which is utilized by 5 teams (Team 6, Team 8, Team 14, Team 16, Team 20), where features with zero variance across all samples are eliminated. This is a significant step to get rid of unnecessary computation during the learner training. Redundant Feature Elimination (RFE) was used by 3 teams (Team 15, Team 16, Team 20) which a similar objective to Constant Feature Elimination. Basically, RFE removes unnecessary features such as duplicates or all-zero features since they do not contribute to the learning at all but only increase the computational load. Teams 16 and 20 used both Constant Feature Elimination and Redundant Feature Elimination. 

Three teams (4, 10 and 17) have scaled their data using a standard scaler or a maximum absolute scaler in order to equalize the range among features. Maximum absolute scaler scales each feature by the maximum of feature values so that the maximum value for each feature will be 1. Standard scaler shifts the feature distribution into a normal one where the mean is 0 and variance is 1. Scaling the data forces the features to have similar ranges for the model to learn from each feature adequately without any dominating features. Both Teams 4 and 10 used the same sequence of pre-processing techniques but their pipelines differ in the dimensionality reduction and learning model stages, which explains the difference between their rankings in \textbf{Table}~ \ref{tab:method_component}. Two teams (7 and 19) applied the inverse sigmoid  function to the data assuming the data lays on a sigmoid-like distribution. So, the features within the range of (0, 1) are mapped into $(-\inf, +\inf)$ interval prior to the training and mapped back into the (0, 1) interval using sigmoid function following the prediction step. Teams 7 and  19 constructed their pipelines with the same components but since the parameters of the components are not identical for both teams, it is expected to observe a slight difference in their performance as \textbf{Table}~ \ref{tab:method_component} indicates.

Add Noise method used by Team 18 is an augmentation method that produce more training samples in order to overcome the curse of dimensionality in machine learning tasks \citep{holmstrom1992}. By adding Gaussian noise to the samples, new samples are generated. Both original and synthetic samples are then used for model training. Even if the old and new samples are expected to be highly correlated, this method slightly increased the performance of Team 18. Only Team 17 benefited from correlation-based feature elimination where they calculated the pair-wise correlation between each feature and dropped the features with high correlations. Highly correlated features are kind of linearly dependent and using them together does not give additive valuable information. Team 17 used correlation-based feature elimination together with Outlier Elimination and scaler. When comparing the rankings of Team 17 and Team 14 (\textbf{Table}~\ref{tab:method_component}), it can be concluded that the pre-processing techniques of Team 17 worked better as their performance surpasses that of Team 14 with the same learning model. This shows how significant is to choose the most appropriate pre-processing techniques while designing machine learning pipelines.

\subsubsection{Dimensionality Reduction Methods}

\begin{figure}[h!]
    \centering
    \includegraphics[width=0.9\textwidth]{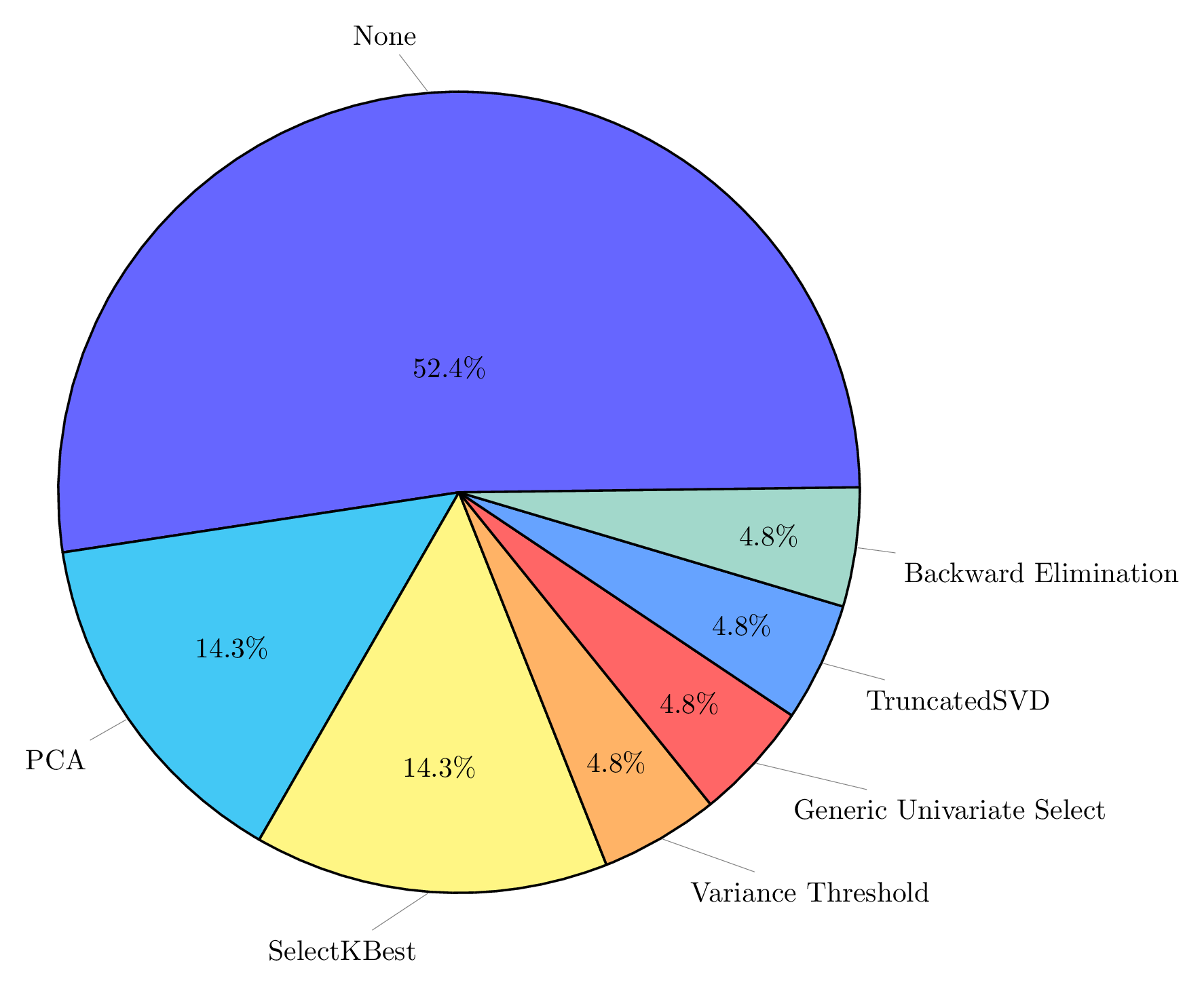}
    \caption{Distribution of the adopted dimensionality reduction techniques among the 20 competing teams.}
    \label{fig:dim_red}
\end{figure}

The curse of dimensionality is one of the well-known issues in the machine learning field where learning from a small sample size with high-feature dimensionality can load the model with unnecessary features, inhibit model generalizability to unseen samples, and increase the computation time for training. Many teams handled this issue using dimensionality reduction techniques which eliminate irrelevant features (feature selection) or extract compact feature representations of the input data (feature extraction). Eventually, a small feature set is obtained with lower-dimensionality. The distribution of the utilized dimensionality reduction techniques among the 20 competing teams is shown in \textbf{Fig.} \ref{fig:dim_red}.

More than half of the competing teams decided not to use any dimensionality reduction technique and passed the feature set into models as it is or only eliminated constant or redundant features. Most adopted techniques are principle component analysis (PCA) (used by Teams 3, 10 and 20) and Select $K$ Best (used by Teams 7,  8 and 19). The remaining ones were used by 1 team each, which are Variance Threshold (Team 4), Generic Univariate Select (Team 3), Truncated SVC (Team 9) and Backward Elimination (Team 17).

\textbf{Feature Selection Methods. } Feature selection methods can handle the high-dimensionality problem by discarding some of the features which have less contribution to the target learning task. The feature selection methods adopted in the competition are Select $K$ Best, Variance Threshold, Generic Univariate Select and Backward Elimination. 

Select $K$ Best algorithm is a supervised feature selection method which applies statistical scoring functions to check the relevancy between the features with respect to the target output data, then selects the least dependent $K$ features in the feature set. The teams measured the dependency between features with mutual information \citep{ross2014mutual}. Team 7 and 19 used Select $K$ Best together with Inverse Sigmoid Function and they adopted Support Vector Regressor \citep{smola2004tutorial} as the learner same with Team 20. When we compare the performance of both teams we note that choosing the most appropriate pre-processing and dimensionality reduction methods was critical in their case. While Team 20 reduced the feature set size using multiple techniques such as CFE, RFE and PCA, Team 7 and 19 applied only one dimensionality reduction method as Select $K$ Best and performed better than Team 20. 
%Team 8 used Select K Best after Constant Feature Elimination and used multiple learners together with Voting Regressor. 

Regarding Variance Threshold, the variance of each feature along the samples is calculated and the features with variances above the determined threshold are chosen for the final feature set. Ultimately, the low-variance features are eliminated which are expected to provide less impact on the regression model learning. Team 4 adopted this method and removed 6 features with lowest variances. Generic Univariate Select provides a system where it opts for the best univariate selection method among Select $K$ Best and Select Percentile where they both measure the dependency between features using statistical analysis and select the least dependent features based on the given number $K$ or a percentage respectively. The optimum parameters are defined by the hyper-parameter estimator and the best resulting univariate feature selection method is elected. Team 3 used this method together with PCA in the dimensionality reduction phase.

Backward Feature Elimination is a supervised feature selection method where some of the features are eliminated based on the evaluation of the training feature set using a learner and training target values (i.e., connectivity features at $t_1$). In each iteration, the least significant features that do not decrease the model performance when removed, are eliminated based on the statistical analysis results. Such method provides an accurate information regarding the importance of the features since it directly evaluates the worth of each feature based on the target learning task. Team 17 adopted this method by training a least square regressor on the input feature set and eliminating the least necessary features in each step.

\textbf{Feature Extraction Methods. } Feature extraction methods handle the high-dimensionality problem by generating a more compact set of features. They differ from the feature selection methods by learning new feature representations. The feature extraction methods used in this competition are PCA by 3 teams and Truncated singular value decomposition (SVD) by 1 team.

PCA \citep{minka2000automatic} is a common feature extraction method where the feature set is projected into a lower-dimensional space in a fully unsupervised manner. When deciding the projection hyperplane, the eigenvalues of the covariance matrix of the data feature are calculated and the eigenvectors with the highest eigenvalues define the new projection space. So that, the dimension that captures the highest data variance is used for the projection by minimizing the loss of information and achieving a more compact representation of the feature set. Truncated SVD \citep{manning2008matrix} maps the high-dimensional feature set into a lower dimension using singular value decomposition similar to PCA. However in PCA, the data is centered by the per-feature mean value while there is no centering step for Truncated SVD.

Team 3 reduced the size of feature set into 21 using PCA followed by one more dimensionality reduction step with Select $K$ Best. Given that the difference between the original feature set size and new feature set size is extremely high, Team 3 still ranked within the top 5 as \textbf{Table}~ \ref{tab:method_component} shows. Thus, we can remark that it is possible to successfully represent the given feature set in a much lower dimension capturing the essential information about the input data. However, we note that the choice of the regression model remains as a significant factor for the overall ML framework performance as other teams which employs PCA or Truncated SVD did not perform as good as Team 3. Furthermore, comparing the ranks of Team 5 and 10, despite the fact that they both utilized the same type of learner and their pre-processing steps do not differ much, Team 10 using PCA ranked much lower than Team 5 which did not use any dimensionality reduction technique. Meaning that reducing the dimension of data may negatively affect the overall model performance in some cases depending on the pipeline and dimensionality reduction parameters. Such choices need to be handled with utmost care and rigor.

\subsubsection{Learning Models}

\begin{figure}[h!]
    \centering
    \includegraphics[width=0.9\textwidth]{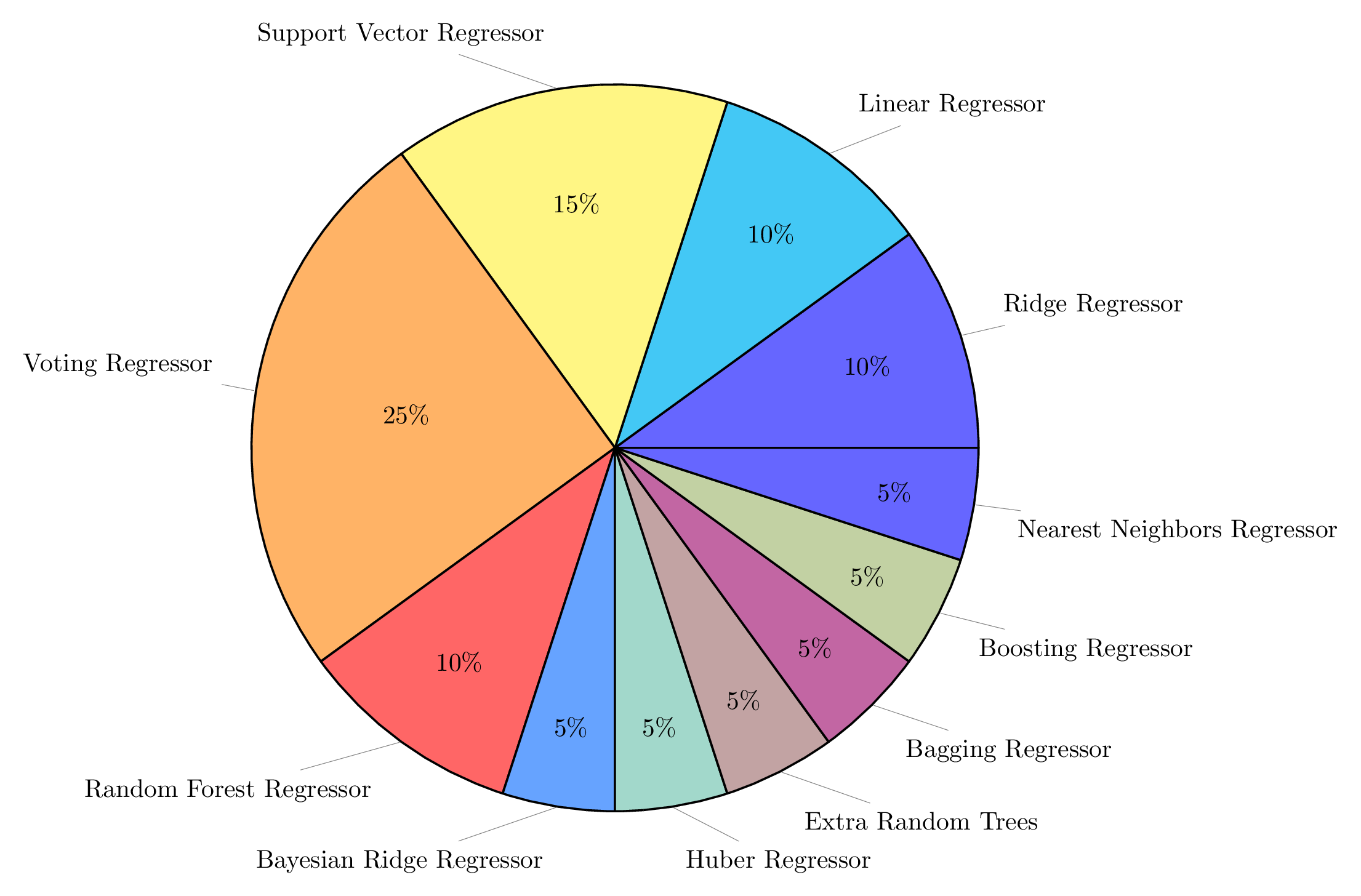}
    \caption{Distribution of the used learning models among the 20 competing teams.}
    \label{fig:learner}
\end{figure}

Following the data preprocessing and dimensionality reduction steps, we reach the final block of the ML pipeline which is the adopted learning model. The trained learner should be able to predict the brain connectivity weights at a follow-up timepoint $t_1$ given the brain connectome at $t_0$ in the form of a matrix. Such a problem can be formalized as a regression problem. Teams participated in the competition have adopted a large variety of \emph{regression models} and a many competitors used ensemble models (voting regressor, bagging regressor and boosting regressor) which rely on more than one learner to predict the target brain connectome at $t_1$. The distribution of the used learning models among the 20 competing teams is shown in \textbf{Fig.}~\ref{fig:learner}.

Five teams (1, 2, 3, 11 and 12) used feature focused learning (FFL) which makes it possible to train models separately for each feature instead of training a single learner on the whole feature set. A learning model is trained independently for each feature (i.e., brain connectivity weight). Next, the feature-wise prediction results are combined into one sample (i.e., the brain connectivity map). While this maximizes the information extracted from a single feature, it does not benefit from the correlations between features. However, we notice that the teams which used feature focused learning ranked in top 10 (\textbf{Table}~\ref{tab:method_component}) and this indicates that such a feature-wise learning strategy can successfully work on connectomic datasets. In addition, comparing the pipelines of Team 11 and 13, they both used a single outlier elimination method and used the Linear Regressor as a learning model. The most obvious difference between both teams is that Team 11 used FFL and performed slightly better while Team 13 trained the model on the whole feature set. When using feature focused learning, at least one base regressor is needed since FFL is not a learning model itself but a training strategy. Competing teams exploited various base learners for FFL where most of them used a single learner and Team 3 used an ensemble method together with FFL. In addition, all teams excluding 3 used the whole feature set during training as they did not apply any dimensionality reduction, thus the number of models that are trained is equal to the number of features. As for the Team 3, they have reduced the size of the feature set to minimize the number of models trained.

Linear Regressor (used by Team 4, Team 11 (FFL), Team 13) is one of the most basic regression models, aiming to fit a linear model with estimated weights, or coefficients. The weights are estimated by minimizing the sum of the squares between predicted outputs based on the estimated weight parameters and the actual outputs. To avoid overfitting and better model the input-output data relationship, a regularization term is usually added to the regression loss function to optimize.   %There are various versions of Linear Regression models based on the constraint that is put on the weights referred to as regularization. Regularization is used to penalize the weight values that surpass a defined threshold and prevents the model to overfit. 
There are 4 other types of Linear Regressors used in the competition:
\begin{itemize}
    \item Orthogonal Matching Pursuit (OMP): Linear Regressor with L0 regularization. L0 regularization penalizes the weight vector based on the number of non-zero parameters. (Team 3)
    \item Lasso: Linear Regressor with L1 regularization. L1 regularization penalizes the weight vector based on the L1 norm and prevents the L1 norm of the weight to be higher than a threshold. (Team 3) 
    \item Ridge Regression: Linear Regressor with L2 reguarization. L2 regularization is similar to the L1 regularization using the L2 norm instead of L1. (Team 3, Team 4, Team 14, Team 17)
    \item Elastic Net: Elastic Net includes a regularization term combining L1 and L2 regularization. (Team 3)
\end{itemize}

Note that Teams 3 and 4 used Linear Regressor and its variants together with other learning models through Voting Regressor while Teams 11 and 13 used only the Linear Regressor as a learner. It is worth to mention that Team 13 constructed a rather simple pipeline with only outlier elimination and Linear Regression still ranked higher than most of the teams with more complex learning phases including Team 4 which used Linear Regressor together with a Voting Regressor.

Team 1 employed FFL using the Bayesian Ridge Regressor as the base model and Team 15 also included it in their set of models for Voting Regressor. Bayesian Ridge Regressor \citep{mackay1992bayesian} is a supervised regression model where a Bayesian approach is applied to a Ridge Regressor. Applying Bayesian to Ridge Regressor provides us with probabilistic model where the targets are assumed as sampled from a normal distribution and the model parameters are determined using the posterior distribution given the prior one. Even though Teams 8 and 15 utilized similar pre-processing techniques and learners, Team 15 ranked higher as they included the Bayesian Ridge Regressor in their set of regression models.

Huber Regressor (used by Team 2) is a regression model robust to the outliers as stated in \citep{owen2007robust}. It calculates linear loss for outlier samples different from the regular samples where their loss values are calculated with a quadratic function. Thus, one can reduce the negative effect of outliers on the learning while making use of these samples as well instead of completely ignoring them. Team 2 used Huber Regressor with FFL and did not apply any pre-processing on data, still ranked 2nd as \textbf{Table}~\ref{tab:method_component} shows. %This indicates that using an outlier insensitive learning model without any outlier elimination gives satisfying performance on this task. 
 Also Team 2 ranked higher than Team 12 although they share the same ML pipeline excluding the learner type, showing the advantage of Huber Regressor in the target learning task.

Support Vector Regressor (used by Teams 7, 19 and 20) \citep{smola2004tutorial} transforms the regular support vector machine classifier to a regression model. Different from the linear regression, SVR can map the input data into both linear and non-linear spaces in order to get a better representation of the data for the target regression task. %The loss for SVR is based on the margin loss where the loss value of the sample is added to the overall loss only if the difference between its estimated output and the ground truth is above a threshold. Meaning that the effect of the correctly predicted outputs on the loss is less than the incorrect predictions. % 
Teams 7 and 19 used the same ML pipelines including the SVR and pre-processing parts, ranking in consecutive places. When comparing the ranks of Teams 7 and 19 with Team 8, all using identical pre-processing steps, we notice that using a single SVR model worked better than the ensemble methods used by Team 8 (\textbf{Table}~\ref{tab:method_component}).

K Nearest Neighbors Regression (used by Teams 3, 8, 12 and 15) proposed by \citep{altman1992introduction} is a non-parametric model where no parameters are determined prior to the prediction step. Predictions are made depending on the outputs of k-nearest neighbors of a given testing sample. The nearest neighbors are specified by calculating the distances between samples using various distance functions. After finding the k nearest  neighbors, the output (i.e., follow-up connectome) of a given testing sample is predicted as the combination of ground truth outputs of the k nearest neighbors in the training set.% Here, combining the outputs can be done under various constraints such as uniformly weighting the each neighbor's output or defining weights based on the distances. %
Teams 3, 8 and  15 used K Nearest Neighbors Regression as one of the base regressors for Voting Regressor while Team 12 used Singular Neighbors Regression as a single learner. Here, we can address the superiority of Singular Neighbors Regression over Extra Random Trees as Team 12 ranked higher than Team 18 with a similar pipeline as shown in \textbf{Table}~\ref{tab:method_component}.

A few teams exploited ensemble methods using a set of the aforementioned methods as base models. The motivation for using ensemble methods is that joining forces and merging expertise from multiple models can improve the prediction results. The adopted ensemble methods can be listed under two categories: averaging methods and boosting methods.

\emph{Averaging ensemble methods}. Random Forest Regressor, Extra Random Trees Regressor, Voting Regressor and Bagging Regressor were used to train multiple models independent from each other and take the average of their  results to produce the final follow-up connectivity map.

Random Forest Regressor (used by Team 5, Team 9 Team 10) trains multiple Decision Trees on random sample subsets and combines their outputs to get the final prediction for test samples \citep{breiman2001random}. Team 9 used Random Forest Regressor together with K-Means Clustering \citep{sculley2010web} where they clustered the training samples to create the dataset splits for Random Forest instead of randomly splitting it into subsets. Team 18 utilized Extra Random Trees \citep{geurts2006extremely} which gets one step further in terms of randomness and decision tree feature splits are determined randomly instead of choosing the threshold that gives the most discriminative feature splits. 

As for the Voting Regressor (used by Teams 3,  4,  8,  9 and 15), different types of models were trained on the connectomic dataset and all of them produced the final brain connectivity predictions by averaging their outputs. Notably, it is possible to take the weighted average as well, defining different weights for each model output. In the competition, various combinations of regression models were used together with Voting Regressor as \textbf{Table}~\ref{tab:method_component} shows. Using Voting Regressor enhances the contributions of the best performing models to the target prediction task, thereby benefitting from their diverse strengths. Team 3 used a wide variety of regression models and largely benefitted from using the Voting Regressor as it ranked in the top 5 teams. However, the selection of the base regressors for Voting Regressor is crucial since other teams using such a strategy failed to perform as good as the Team 3.

\emph{Boosting ensemble methods.} Bagging Regressor \citep{louppe2012ensembles} takes a base regressor and creates several instances from this model type. These models are then trained using different splits of the data samples independently. Next, the different trained models are separately used to predict the output for given testing sample and all predictions are combined into one final prediction. Team 6 employed the Bagging Regressor choosing Partial Least Squares Regression as the base model which attempts to infer latent variables from the data, then used those variables to discover a linear relationship within the data as stated in \citep{wegelin2000survey}. However, when comparing the performances of Teams 6 and 1 which have similar pre-processing steps, one notices that Team 1 has outperformed Team 6 with a large gap using Bayesian Ridge Regressor with FFL highlighting its advantage for this specific task.

Boosting Regressors utilized in the competition are Gradient Boosting Regressor, XGradient Boosting Regressor, Light Gradient Boosting Regressor and AdaBoost Regressor which dependently train multiple weak models where each learner aims to improve the weaknesses of its predecessor learner \citep{drucker1997improving}. XGradient Boosting and Light Gradient Boosting Regressors (used by Team 3) are variants of Gradient Boosting Regressor  \citep{friedman2001greedy} which learns from the errors of the previous models in each step and then forces the next model to decrease the overall error by taking steps in the direction of the negative gradient of the defined loss function similar to the gradient descent strategy. AdaBoost Regressor (used by Team 3, Team 8, Team 15, Team 16) also aims to decrease the overall loss in each step based on the errors in the previous step. In order to do that, it gives more weight to samples with high error values while building the next model. As shown in \textbf{Table}~\ref{tab:method_component}, only 3 teams ranked within the top 10 using ensemble methods while most of the highly ranked teams preferred simpler methods indicating that the efficacy of ensemble methods for this task is arguable.

%%%%%%%%%%%%%%%%%%%%%%%%
%%%%%%%%%%%%%%%%%%%%%%%%
\section{Results}

\subsection{Evaluation Dataset}  

We used $230$ subjects from the OASIS-2\footnote{\url{https://www.oasis-brains.org/}} longitudinal dataset \citep{Marcus:2010}. This set consists of a longitudinal collection of 230 subjects aged 60 to 96. Each subject has 3 visits (i.e., timepoints), separated by at least one year. For each subject, we construct a cortical morphological network derived from the cortical thickness measure using structural T1-w MRI as proposed in \citep{mahjoub2018,corps2019,nebli2020}. To this aim, we first use FreeSurfer \citep{fischl2012freesurfer} to construct the cortical surface for each hemisphere, which is then parcellated into $35$ ROIs using Desikan-Killiany cortical atlas \citep{desikan2006automated}.  Next, we compute the connectivity weight between two ROIs (i.e., nodes) by compute the absolute difference in their cortical thickness. Such cortical morphological networks quantifying the dissimilarity in morphology between cortical regions provided novel insights into dementia \citep{mahjoub2018,lisowska2019joint,raeper2018} and autism \citep{soussia2018} diagnosis, brain age prediction \citep{corps2019}, gender differences \citep{nebli2020} and infant brain growth fingerprinting \citep{rekik2018baby}.

Similar to the brain network classification competition \citep{BILGEN2020108799}, the competitors are given a train set of 150 sample. For evaluation, they can test their models on a public test set including 40 samples, while another private test set with 40 samples is kept hidden throughout the competition. Competitors could access the results of their ML pipelines on the public test set immediately in order to check the error score of their models, and the performance of their final submitted models are evaluated on a private set after the competition has ended. This way, competitors are encouraged to prioritize generalizability of their models in order to score high at the end of the competition. 

\subsection{Performance Evaluation} 

%need help(D)

\begin{table}[]
\centering

  \begin{adjustbox}{width=\textwidth}
\begin{tabular}{|>{\columncolor[RGB]{247,231,206}}c|c|c|c|c|c|c|>{\columncolor[RGB]{252,240,240}}c|c|c|c|c|c|c|>{\columncolor[RGB]{254,249,237}}c|>{\columncolor[RGB]{251, 236, 150}}c| l p{2cm} }
\hline

\multirow{1}{*}{ \cellcolor[RGB]{247,231,206}} & \multicolumn{7}{c|}{ \cellcolor{babypink!25}\textbf{MAE}}   & \multicolumn{7}{c|}{ \cellcolor{bananamania!25}\textbf{PCC}}                                                                      &  \\ \cline{2-15}
                         \multirow{1}{*}{ \cellcolor[RGB]{247,231,206}\textbf{Team}} & \multicolumn{2}{c|}{ \cellcolor{babypink!25}Public} & \multicolumn{2}{c|}{ \cellcolor{babypink!25}Private} & \multicolumn{2}{c|}{ \cellcolor{babypink!25}5 CV} &  \cellcolor{babypink!25}Rank & \multicolumn{2}{c|}{ \cellcolor{bananamania!25}Public} & \multicolumn{2}{c|}{ \cellcolor{bananamania!25}Private} & 
                         \multicolumn{2}{c|}{ \cellcolor{bananamania!25}5 CV} &  \cellcolor{bananamania!25}Rank &  
                         \textbf{Final Rank}
                          \\ \hline
1                       & 0.0309   & 1       & 0.0343      & 1        & 0.0363     & 2      & 1    & 0.797      & 1        & 0.743       & 2         & 0.687     & 6       & 3    & \textbf{1}                           \\ 
2                       & 0.0310   & 2       & 0.0344      & 2        & 0.0362     & 1      & 2    & 0.795      & 3        & 0.741       & 3         & 0.691     & 2       & 2    & \textbf{1}                           \\ 
11                      & 0.0311   & 3       & 0.0344      & 3        & 0.0363     & 3      & 3    & 0.797      & 2        & 0.743       & 1         & 0.691     & 3       & 1    & \textbf{1}                            \\
13                      & 0.0314   & 4       & 0.0348      & 4        & 0.0366     & 4      & 4    & 0.793      & 4        & 0.739       & 4         & 0.690     & 4       & 4    & \textbf{4}                            \\ 
3                       & 0.0320   & 8       & 0.0353      & 6        & 0.0368     & 5      & 6    & 0.787      & 6        & 0.737       & 5         & 0.698     & 1       & 4    & \textbf{5}                           \\
12                      & 0.0317   & 5       & 0.0351      & 5        & 0.0371     & 6      & 5    & 0.792      & 5        & 0.736       & 6         & 0.688     & 5       & 6    & \textbf{6}                            \\ 
7                       & 0.0320   & 6       & 0.0356      & 8        & 0.0372     & 8      & 8    & 0.782      & 8        & 0.732       & 8         & 0.683     & 10      & 7    & \textbf{7}                         \\ 
19                      & 0.0320   & 7       & 0.0356      & 7        & 0.0372     & 7      & 7    & 0.781      & 9        & 0.731       & 9         & 0.683     & 11      & 10   & \textbf{8}                          \\ 
18                      & 0.0322   & 9       & 0.0358      & 11       & 0.0376     & 11     & 10   & 0.784      & 7        & 0.727       & 13        & 0.684     & 8       & 8    & \textbf{9}                          \\ 
16                      & 0.0325   & 11      & 0.0358      & 12       & 0.0373     & 10     & 11   & 0.778      & 14       & 0.734       & 7         & 0.687     & 7       & 8    & \textbf{10}                         \\ 
4                       & 0.0323   & 10      & 0.0357      & 9        & 0.0373     & 9      & 9    & 0.780      & 12       & 0.730       & 11        & 0.684     & 9       & 11   & \textbf{11}                        \\
15                      & 0.0325   & 12      & 0.0357      & 10       & 0.0377     & 13     & 12   & 0.780      & 13       & 0.731       & 10        & 0.680     & 14      & 12   & \textbf{12}                         \\ 
5                       & 0.0326   & 13      & 0.0361      & 14       & 0.0378     & 14     & 13   & 0.781      & 10       & 0.724       & 14        & 0.676     & 15      & 13   & \textbf{13}                        \\ 
17                      & 0.0327   & 14      & 0.0360      & 13       & 0.0391     & 18     & 15   & 0.780      & 11       & 0.729       & 12        & 0.668     & 17      & 14   & \textbf{14}                         \\ 
20                      & 0.0333   & 16      & 0.0365      & 16       & 0.0377     & 12     & 14   & 0.772      & 16       & 0.718       & 19        & 0.682     & 12      & 15   & \textbf{14}                         \\ 
8                       & 0.0338   & 19      & 0.0366      & 17       & 0.0384     & 17     & 17   & 0.759      & 20       & 0.722       & 15        & 0.681     & 13      & 16   & \textbf{16}                        \\ 
14                      & 0.0330   & 15      & 0.0362      & 15       & 0.0380     & 16     & 16   & 0.770      & 17       & 0.720       & 17        & 0.667     & 18      & 18   & \textbf{17}                         \\
9                       & 0.0336   & 18      & 0.0369      & 20       & 0.0379     & 15     & 17   & 0.764      & 18       & 0.721       & 16        & 0.671     & 16      & 17   & \textbf{17}                        \\
6                       & 0.0333   & 17      & 0.0369      & 19       & 0.0405     & 20     & 19   & 0.773      & 15       & 0.717       & 20        & 0.652     & 19      & 19   & \textbf{19}                        \\ 
10                      & 0.0339   & 20      & 0.0366      & 18       & 0.0398     & 19     & 20   & 0.759      & 19       & 0.719       & 18        & 0.642     & 20      & 20   & \textbf{20}                         \\ \hline
\end{tabular}
  \end{adjustbox}
\caption{\label{tab:rankings} \emph{Final team rankings.} Each column displays the evaluation metric value (left) and the team rank (right). The final rank is determined based on the local ranks using the public, private and 5 CV scores across both MAE and PCC evaluation metrics. MAE: mean absolute error. PCC: Pearson correlation coefficient. 5 CV: 5-fold cross-validation. Public: performance on the public test set. Private: performance on the private test set.}

\end{table}

\emph{Evaluation metrics.} For model evaluation, we use the Mean-Squared Error (MSE), Mean-Average Error(MAE) and Pearson Correlation Coefficient (PCC) on both public and private test sets. Competitors had access to baseline brain connectivity matrices only at time step $t_0$ and can only check the MSE score of their models on the public test subset, whereas MSE scores on the private subset are revealed after the competition ended to prevent the overfitting to the test set, thus, ensuring the generalizability of the best performing Ml pipelines. Since MAE is more robust to outliers than the MSE, each team is ranked based on two evaluation measures: MAE and PCC scores. While MAE captures the large differences between the predicted brain connectivity at follow-up timepoint $t_1$ and the ground truth one (closer to 0 indicates better prediction), the PCC captures their correlation (closer to 1 implies better prediction). Hence, both measures allow us to evaluate different aspects of the learned models. 

\emph{Final ranking.} For each evaluation measure (MAE and PCC), teams are assigned 3 rankings according to their public, private and 5-CV scores. Next, we produce a measure-based ranking for each team by averaging their 3 local rankings. The final rankings are then computed by taking the average of MAE and PCC ranks as displayed in \textbf{Table}~\ref{tab:rankings}.

\begin{figure}[h!]
    \centering
	\includegraphics[width=14cm]{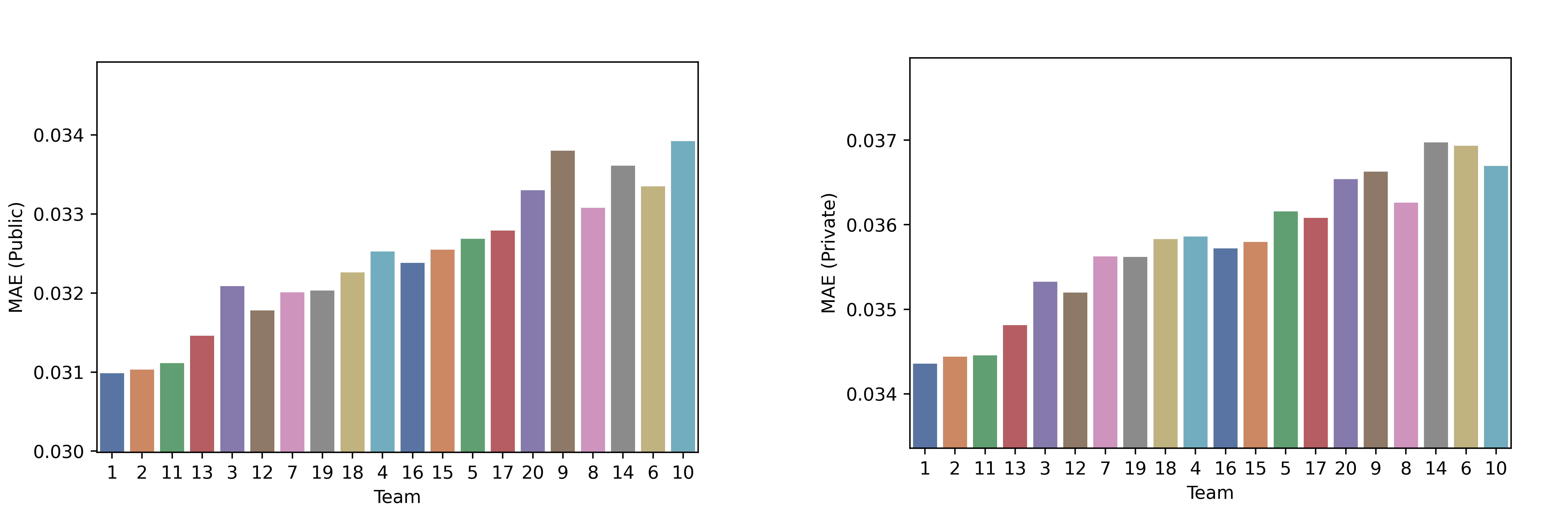}
    \caption{Performance of the 20 competing teams using the Mean Average Error (MAE). Left: MAE scores on the public test set. Right: MAE scores on the private test set.}
    \label{fig:mae_barplot}
\end{figure}

\begin{figure}
    \centering
    \includegraphics[width=14cm]{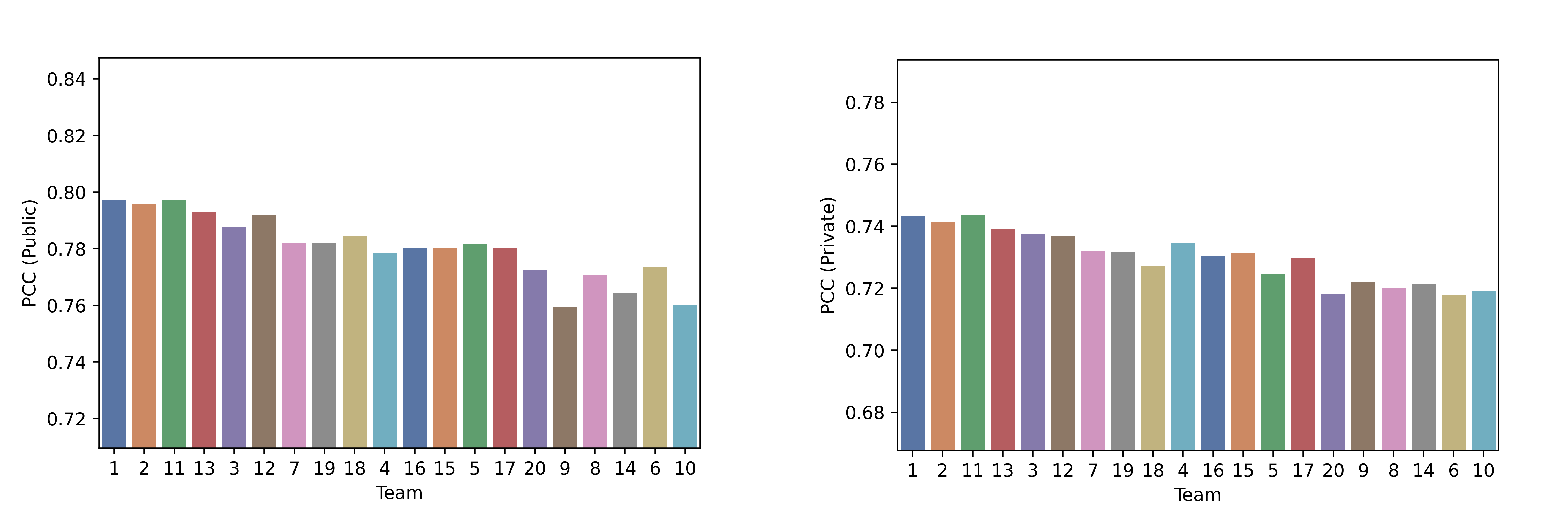}
    \caption{Performance of the 20 competing teams using the Pearson Correlation Coefficient (PCC). Left: PCC scores on the public test set. Right: PCC scores on the private test set.}
    \label{fig:pcc_barplot}
\end{figure}

\emph{Overall performance.} \textbf{Table}~\ref{tab:method_component} displays the components of the ML pipeline designed by each competitor, while the performance of the competing teams and their rankings are summarized in \textbf{Table}~\ref{tab:rankings}. The PCC results are given in \textbf{Fig.}~\ref{fig:pcc_barplot} and the MAE results in \textbf{Fig.}~\ref{fig:mae_barplot}. The teams are ordered with respect to their final rank. Given the variety of preprocessing, dimensionality reduction and learning methods within the competitors, we could observe results from diverse learning pipelines. 

\begin{figure}[h!]
    \centering
    \includegraphics[width=14cm]{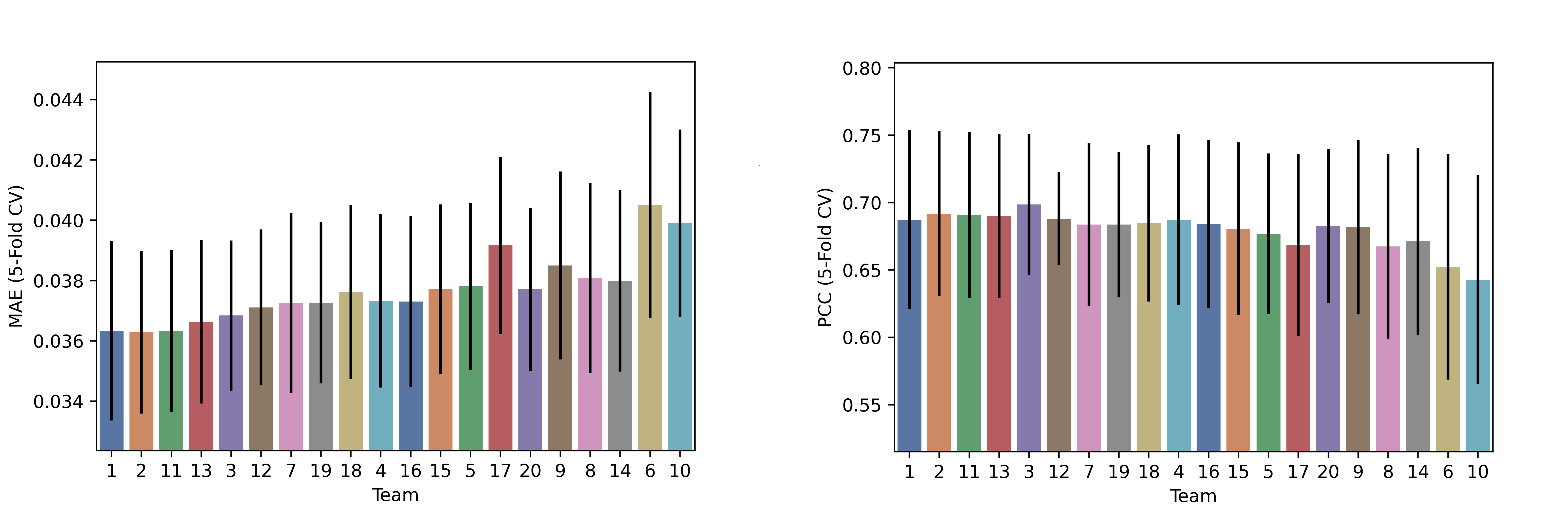}
    \caption{Average MAE and PCC results of the 20 competing teams using 5-fold cross-validation. The vertical bar denotes the standard deviation of the evaluation score across 5 folds.}
    \label{fig:5fold_barplots}
\end{figure}

The competing teams used 10 different preprocessing methods and 6 different dimensionality reduction methods in total. In terms of preprocessing methods, the most notable outcome is that teams with multiple preprocessing methods performed worse than other teams overall, while there is no major difference in terms of performance between teams with 1 data preprocessing method and no preprocessing. In terms of dimensionality reduction methods, the best performing teams opted not to use any dimensionality reduction technique, while PCA and Select K-Best methods were the most common methods among competitors.

\begin{figure}[h!]
    \centering
    \includegraphics[width=14cm]{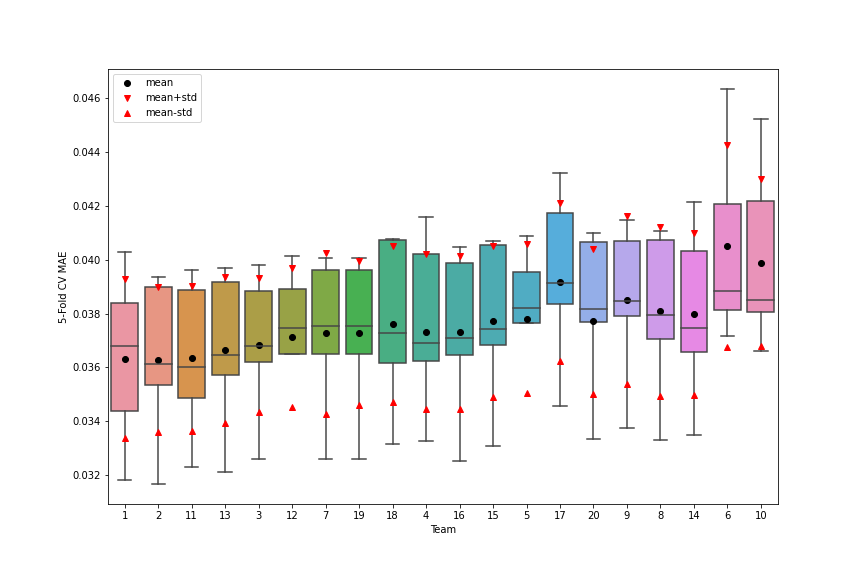}
    \caption{The Mean Average Error (MAE) of the 20 competing teams using using 5-fold cross-validation.}
    \label{fig:mae_boxplot}
\end{figure}

\begin{figure}[h!]
    \centering
    \includegraphics[width=15cm]{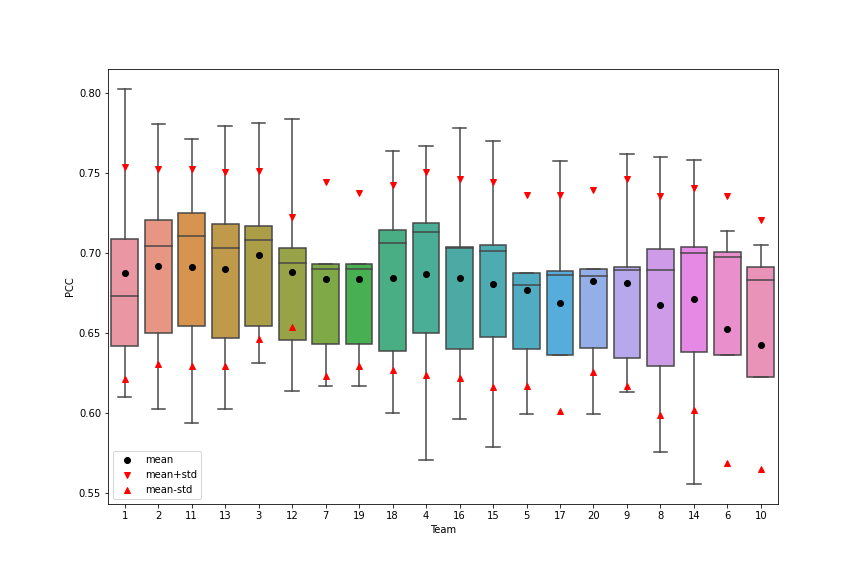}
    \caption{The Pearson Correlation Coefficient(PCC) of the competing teams using 5-fold cross-validation.}
    \label{fig:pcc_boxplot}
\end{figure}

\begin{figure}[h!]
    \centering
    \includegraphics[width=14cm]{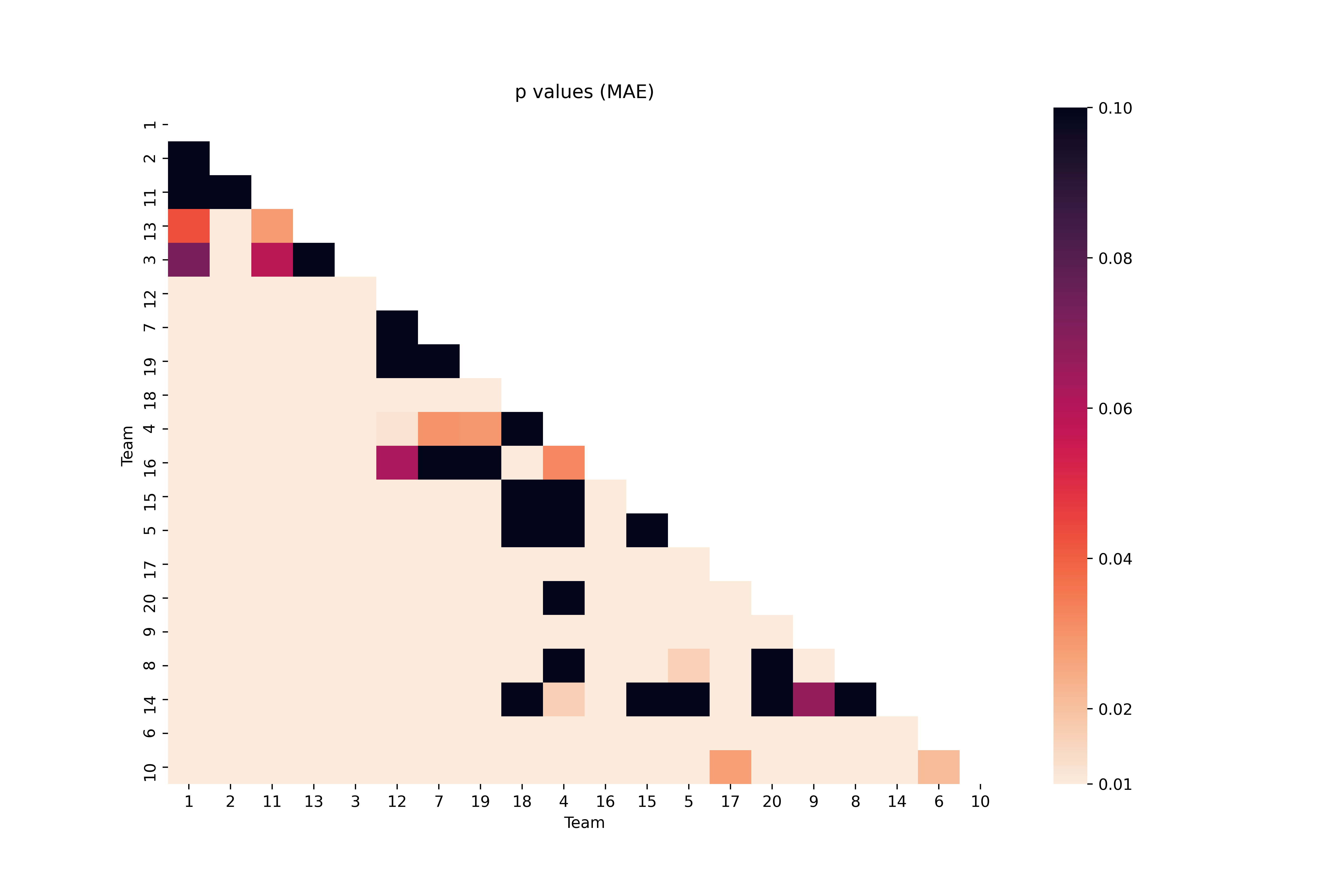}
    \caption{p-values using two tailed paired t-test on the MAE evaluation scores of the 20 competing teams.}
    \label{fig:mae_heatmap}
\end{figure}

\begin{figure}[h!]
    \centering
    \includegraphics[width=14cm]{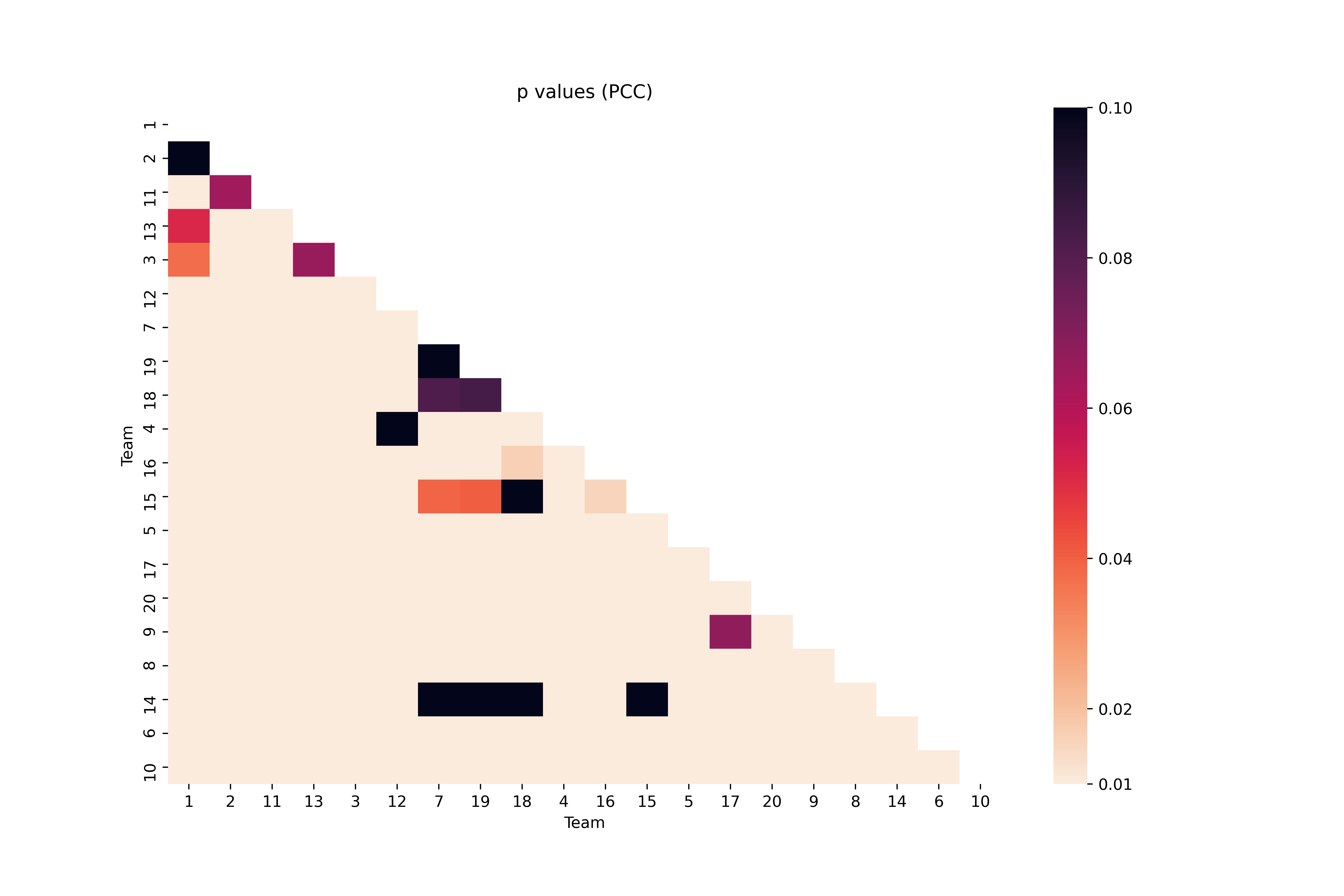}
    \caption{p-values using two tailed paired t-test on the PCC evaluation scores of the 20 competing teams.}
    \label{fig:pcc_heatmap}
\end{figure}

The learning methods of the best performing teams included Bayesian Ridge Regressor(BR) (Team 1 ranked first),  Huber Regressor (Team 2) and Linear Regression (Teams 11 and 13), respectively. A commonly adopted approach by the top performing teams was to utilize Feature Focused Learning (FFL), which was adopted by 5 teams in the top 6, while remaining unexplored in the lower half of the rankings. It should be noted that teams that used FFL also opted to not use any dimensionality reduction method, rather they aimed to predict each brain connectivity weight (i.e., feature) independently using the original data. Interestingly, this reveals that focusing on predicting each feature independently of other features led to the best performing pipelines in this competition. Another distinguishing approach among the teams was to use a single learner or to utilize multiple learners through the Voting Regressor strategy. Teams 1, 2, 11, and 13, which are the top performing ones, opted to use single learners, which shows that using multiple learners do not necessarily provide a visible advantage over single learners --in general. On the other hand, Team 3 opted to use 9 different learning methods, breaking the record of most used learners across all teams, and ranked fifth in the competition. 

\textbf{Fig.}~\ref{fig:mae_barplot} and \ref{fig:pcc_barplot} display in bar plots the performance of the teams using the average MAE and PCC sores on both public and private test sets. It is noticeable that the performance of the teams drops continually in a linear fashion, leading to a small performance gap between consecutive teams, however the difference between the high and low rank teams remains large. Furthermore, \textbf{Fig.}~\ref{fig:5fold_barplots} shows the MAE and PCC results of the 20 competing teams using 5-fold cross-validation. We notice that the errors of the lower ranked teams are higher, showing an inconsistent performance across different folds. This indicates the instability of the designed ML pipelines against training and test data distribution shifts.

\textbf{Fig.}~\ref{fig:mae_boxplot} and \ref{fig:pcc_boxplot} display the performance of the 20 competing teams using both complementary MAE and PCC evaluation metrics in box plots. The results of the majority of the competitors across different folds do not display a symmetric distribution but rather a skewed distribution in a positive or negative direction. Although some of the teams performed in a symmetric fashion in one of the metrics (e.g., Team 1 in MAE score), almost all the remaining teams have a non-symmetric performance across the folds using at least one metric, with Team 4 being an exception, showing a symmetry in both metrics, therefore having consistent performance across all folds. This demonstrates the reproducibility issue of ML pipelines which can be very sensitive to shifts in training and testing data distributions \citep{georges2018data,georges2020}.

\begin{figure}[h!]
    \centering
    \includegraphics[width=14cm]{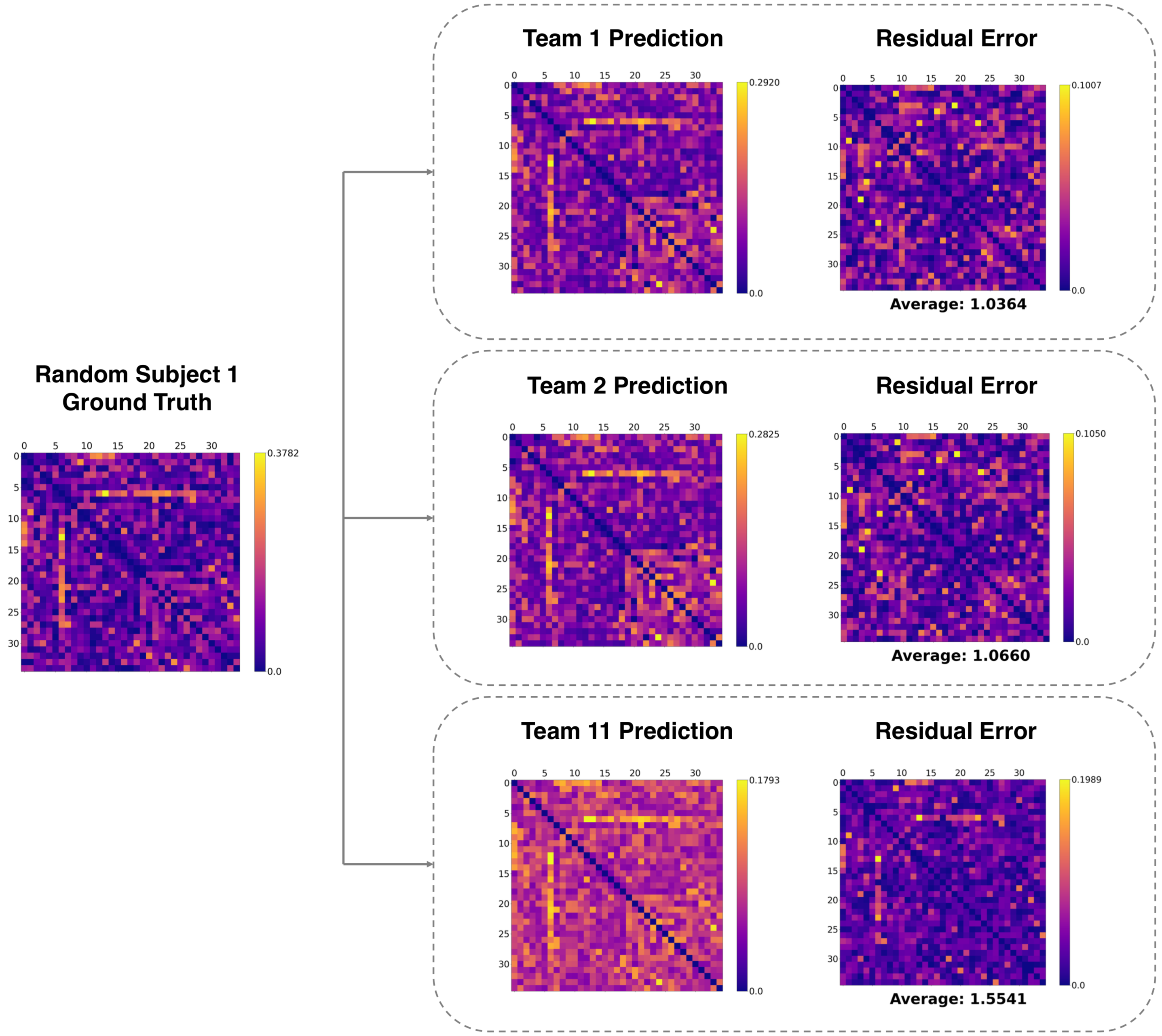}
    \caption{\emph{Predicted morphological brain connectivity matrices at follow-up timepoint by the top 3 performing teams}. We display the predicted brain connectivity matrix at follow-up timepoint and the element-wise absolute difference matrix (i.e., residual) between the ground truth and predicted matrices. We notice that the modular connectome structure is well preserved as reflected by the diagonal blocks.}
    \label{fig:predictions}
\end{figure}

\textbf{Fig.}~\ref{fig:mae_heatmap}  and \ref{fig:pcc_heatmap} show the results of the two-tailed paired t-test p-values between pairs of competing teams using MAE and PCC scores, respectively. Teams on the axes are ordered by their rankings. Teams 1, 2 and 11, who share the first place, got a high p-value on MAE metric when paired, while a high p-value is observed between Team 1 and 2, and between Team 2 and 11 in PCC metric. This shows that the performances of these teams against each other is not very significant and since they share the first place, a further investigation might be needed to determine which of the learning pipelines is more suited to the task in hand in terms of other evaluation criteria such as computational resources. On the other hand, Team 13, which used Linear Regression similarly to Team 11 without benefitting from FFL, ranked fourth right after the first three teams. However, the p-values between Teams 13 and  11 for PCC metric are low $p \sim 0.01$, while they are slightly higher than $0.02$ for MAE metric. This indicates that the performance difference between teams 11 and 13 is statistically significant, demonstrating that the choice of using FFL had an actual impact on the results. As shown in \textbf{Fig.}~\ref{fig:mae_heatmap}  and \ref{fig:pcc_heatmap}, high p-values are mostly observed between teams with a similar performance, meaning the actual rankings between close teams may be swapped, but the performance difference between a high-ranking team and a low-ranking team is statistically significant ($p<0.05$).

\textbf{Fig.}~\ref{fig:predictions} shows the predicted connectivity matrices at follow-up timepoint $t_1$ by the top 3 performing teams (1, 2 and 11) for a representative subject and the residual matrices between the ground truth and predicted brain connectivity matrices, calculated using element-wise absolute difference. Even though the range of the values differs between the teams and the ground truth, it can be observed that the modular structure of the brain connectivity is well preserved in the predictions, indicating that the top performing ML pipelines are capable of foreseeing the change in the brain topological structure over time. 

%%%%%%%%%%%%%%%%%%%%%%%%
%%%%%%%%%%%%%%%%%%%%%%%%
\section{Discussion}
%%%%%%%%%%%%%%%%%%%%%%%%
%%%%%%%%%%%%%%%%%%%%%%%% 

In this study, we have investigated the performance of 20 machine learning pipelines for predicting the evolution of brain connectivity over time given a baseline brain network. To do so, an in-class Kaggle competition was set up for competitors to develop their own learning pipelines and compete with each other, allowing for an investigation of which machine learning methods perform  best on such a challenging problem in the field of \emph{predictive connectomics}. Crowdsourcing using a community competition has been a common approach in a broad range of research areas \citep{BELCASTRO201838,marbach,BRON2015562,saez2016crowdsourcing,BELCASTRO201838,BILGEN2020108799}; however, to the best of our knowledge, this is \emph{the first competition} to tackle the \emph{longitudinal brain connectivity prediction} challenge. 

%In the setup competition, teams were composed of maximum 5 members to develop their learning pipelines. 
Throughout the competition, teams were not granted access to the private test set and were only allowed to evaluate their performance on a public part of the test set by submitting their prediction results generated by their models. Teams were allowed a limited number of result submissions to decrease the likelihood of model overfitting. Upon the completion of the competition, the Kaggle ranking of the participating teams was determined by combining their rankings in both public and private test sets. However, such a ranking strategy is somewhat biased towards the training and testing data samples. To mitigate this issue, the competing teams also provided results of their ML pipelines on the whole dataset. The final rank of each team was then defined as a combination of the 5-fold, public and private test ranks (\textbf{Table}~\ref{tab:rankings}). %In this study, we have only included the top 20 teams out of to investigate the best performing machine learning methods.  

\textbf{Best performing ML pipelines.} In the final rankings shown in \textbf{Table}~ \ref{tab:rankings}, teams are ordered in an ascending order with respect to their final rank combining their ranks using the public and private test sets as well as the 5-fold cross-validation results for both MAE and PCC evaluation metrics. With such a rigorous ranking system, Teams 1, 2 and 11  ranked first. In fact, Team 1 was placed first using the MAE score (0.0309 in public Set, 0.0343 in private set, 0.0363 in 5-CV) and third using the PCC score. Team 2 ranked second using both metrics and Team 11  third using MAE score and first using PCC score. It should be noted that, as shown in \textbf{Fig.}~\ref{fig:mae_heatmap}  and \ref{fig:pcc_heatmap}, the statistical difference in performance between the top 3 performing teams was insignificant since high p-values (p $ >0.05 $) are observed using two-tailed paired t-test, particularly for the MAE evaluation measure. This is a hint that results of these three teams may not be significantly different and other evaluation criteria can be further investigation for clinical translation such as memory and time resources. One the other hand, the three top teams significantly ($p<0.05$) outperformed low-performing teams using both MAE and PCC evaluation metrics. Next, we focus on analyzing their machine learning pipelines.

\textbf{Table}~\ref{tab:method_component} details the methods used by the top 20 teams in the competition. Team 1 opted for using Local Outlier Factor \citep{breunig2000lof} to remove outlier samples and used Bayesian Ridge Regression \citep{mackay1992bayesian} as a predictive learner. Interestingly, Team 2 did not apply any preprocessing or dimensionality reduction to the data, and used Huber Regressor \citep{owen2007robust} as a learning method. Team 11 chose to remove samples that lay outside the inter-quartile range as a preprocessing step, and used Linear Regression as a predictive model. As the top 3 winning teams, Teams 1, 2 and 11 utilized simplistic learning pipelines by not using any dimensionality reduction method where only Teams 1 and 11 used preprocessing methods, which are mainly outlier elimination methods. This shows the importance of removing outliers prior to any learning step. Remarkably, all three teams used the feature-focused learning strategy, which trains a learner for each feature in an independent manner instead of training a single learner using all features. This shows that, instead of aiming to co-learn the interdependency and correlation among features (i.e., brain connectivity weights), predicting how each feature evolves over time, independently of other features, yielded better scores in this competition.  In fact, Team 13 ranked fourth right behind the first ranked three teams and used Linear Regression as Team 11 did, however they did not adopt a feature focused learning strategy. Instead, they trained a single learner for all features, however got ranked lower than Team 11 across all metrics. It should be noted that they also used a different outlier elimination method which could have affected their overall performance. Moreover, competing teams which used FFL (Teams 1, 2, 11, 3, 12 and 18) ranked in the top half, ascertaining  the idea that FFL might be a better choice for the brain connectivity prediction task.

Teams with multiple preprocessing steps, such as Team 10 who ranked twentieth and used Isolation Forest method \citep{liu2008isolation} and scaled the data, have been less successful in the competition.  In addition, the most successful teams with preprocessing methods used methods only to identify and eliminate outliers, such as LOF or Inter-quartile Range. Teams that chose methods for feature elimination, scaling the data or transforming the data are mostly ranked in the bottom half of top 20. One of the most commonly used methods was constant feature elimination, which appeared in five teams' learning pipelines (Teams 16, 20, 8, 14, 6, ordered by their ranking in the competition). We can conclude that it is important to first eliminate sample outliers while keeping the learning pipeline simple and feature-focused rather than sample-focused. 

Furthermore, teams that chose to reduce the dimensionality of their data have been less successful. The top 4 teams (Team 1, 2, 11 and 13) did not use any dimensionality reduction technique thereby preserving the original dimension of the data. This can be an indication that the number of dimensions in the data is not relatively large (595 features in our case) and any loss of information may drop the model performance. However, it should be noted that the  number of teams that used dimensionality reduction methods is not high (9 teams used a total of 6 different methods), therefore other dimensionality reduction methods can be further investigated in a future study. 

\textbf{Generalizability.} One of the formidable challenges faced in machine learning lies in training a model that produces similar good results when presented with test sets drawn from different distributions, i.e., training a \emph{generalizable} model. \textbf{Fig.}~\ref{fig:mae_boxplot} and \ref{fig:pcc_boxplot} show the MAE and PCC results of the teams using 5-fold cross-validation strategy and their standard deviation across folds. In the MAE boxplot, most of the teams have long lower whiskers, while Teams 6 and 10 have a long upper whisker, indicating that the MAE score across folds largely varies. A similar trend can also be observed for the PCC metric as most teams display a long upper or lower whisker. Model performance fluctuation across validation folds points to a weak model generalizability. There are also cases where the upper and lower quartile whiskers are small or non-existent, such as Team 5 results, however in such cases it can be observed that the mean and the median of the data tend to skew towards one direction, suggesting a similar performance across most folds with one fold displaying a large deviation.

%%%%%%%%%%%%%%%%%%%%%%%%
%%%%%%%%%%%%%%%%%%%%%%%%
\section{Conclusion}
%%%%%%%%%%%%%%%%%%%%%%%%
%%%%%%%%%%%%%%%%%%%%%%%%
Foreseeing the evolution of the brain connectivity over time is a formidable challenge to tackle in network neuroscience which can propel the fields of early neurological disorder diagnosis and abnormality detection towards an unprecedented progress. In this study, we are the first to address such a challenge by crowdsourcing a Kaggle competition benchmarking a large number of automated machine learning frameworks designed by 20 competing teams for brain connectivity evolution prediction from a single timepoint. Combining different ranking measures for better benchmarking, the competition resulted in a unique three-way tie for the first place for Teams 1, 2 and 11. All three teams opted for a \emph{feature focused learning}, showing the benefits of this specific approach over the ``single learner on all features'' for this prediction task. Even though the proposed ML frameworks are designed for this competition, they are implemented to be generic and can be used in any regression problem on graphs with same node size. The proposed 20 ML frameworks and the evaluation dataset are accessible at \url{https://github.com/basiralab/Kaggle-BrainNetPrediction-Toolbox}.

%%%%%%%%%%%%%%%%%%%%%%%%%%%%%%%%%%%%%%%%%%%%%%%%%%%%%%%%%%%%%%%%%%%%%%%%%%%%%%%%%%

%%***************************************************************************** %%
\section{Acknowledgements}
%%***************************************************************************** %%

This work was funded by generous grants from the European H2020 Marie Sklodowska-Curie action (grant no. 101003403, \url{http://basira-lab.com/normnets/}) to I.R. and the Scientific and Technological Research Council of Turkey to I.R. under the TUBITAK 2232 Fellowship for Outstanding Researchers (no. 118C288, \url{http://basira-lab.com/reprime/}). However, all scientific contributions made in this project are owned and approved solely by the authors.

\newpage
\bibliography{Kbiblio}
\bibliographystyle{model2-names}

\end{document}